\documentclass[runningheads]{llncs}
\usepackage{graphicx}

\usepackage{tikz}
\usepackage{comment}
\usepackage{graphicx}
\usepackage{amsmath,amssymb} 
\usepackage{color}
\usepackage{booktabs}
\usepackage{multirow}
\newtheorem{thm}{Theorem}
\usepackage{xfrac}
\usepackage[nice]{nicefrac}
\usepackage{wrapfig}
\usepackage{diagbox}
\usepackage{nohyperref}

\usepackage[capitalize]{cleveref}
\crefname{section}{Sec.}{Secs.}
\Crefname{section}{Section}{Sections}
\Crefname{table}{Table}{Tables}
\crefname{table}{Tab.}{Tabs.}

\usepackage[accsupp]{axessibility}  

\begin{document}

\def\mA{{\mathbf{A}}}
\def\mB{{\mathbf{B}}}
\def\mC{{\mathbf{C}}}
\def\mD{{\mathbf{D}}}
\def\mE{{\mathbf{E}}}
\def\mF{{\mathbf{F}}}
\def\mG{{\mathbf{G}}}
\def\mH{{\mathbf{H}}}
\def\mI{{\mathbf{I}}}
\def\mJ{{\mathbf{J}}}
\def\mK{{\mathbf{K}}}
\def\mL{{\mathbf{L}}}
\def\mM{{\mathbf{M}}}
\def\mN{{\mathbf{N}}}
\def\mO{{\mathbf{O}}}
\def\mP{{\mathbf{P}}}
\def\mQ{{\mathbf{Q}}}
\def\mR{{\mathbf{R}}}
\def\mS{{\mathbf{S}}}
\def\mT{{\mathbf{T}}}
\def\mU{{\mathbf{U}}}
\def\mV{{\mathbf{V}}}
\def\mW{{\mathbf{W}}}
\def\mX{{\mathbf{X}}}
\def\mY{{\mathbf{Y}}}
\def\mZ{{\mathbf{Z}}}
\def\mBeta{{\mathbf{\beta}}}
\def\mPhi{{\mathbf{\Phi}}}
\def\mLambda{{\mathbf{\Lambda}}}
\def\mlambda{{\mathbf{\lambda}}}
\def\mSigma{{\mathbf{\Sigma}}}

\pagestyle{headings}
\mainmatter
\def\ECCVSubNumber{5966}  

\title{Batch-efficient EigenDecomposition for Small and Medium Matrices} 

\titlerunning{Batch-efficient EigenDecomposition}
%
\author{Yue Song\orcidID{0000-0003-1573-5643} \and
Nicu Sebe\and 
Wei Wang}
\authorrunning{Y. Song et al.}
%
\institute{DISI, University of Trento, Trento 38123, Italy\\
\email{yue.song@unitn.it}\\
\url{https://github.com/KingJamesSong/BatchED}
}
\maketitle

\begin{abstract}
EigenDecomposition (ED) is at the heart of many computer vision algorithms and applications. One crucial bottleneck limiting its usage is the expensive computation cost, particularly for a mini-batch of matrices in the deep neural networks. In this paper, we propose a QR-based ED method dedicated to the application scenarios of computer vision. Our proposed method performs the ED entirely by batched matrix/vector multiplication, which processes all the matrices simultaneously and thus fully utilizes the power of GPUs. Our technique is based on the explicit QR iterations by Givens rotation with double Wilkinson shifts. With several acceleration techniques, the time complexity of QR iterations is reduced from $O{(}n^5{)}$ to $O{(}n^3{)}$. The numerical test shows that for small and medium batched matrices (\emph{e.g.,} $dim{<}32$) our method can be much faster than the Pytorch SVD function. Experimental results on visual recognition and image generation demonstrate that our methods also achieve competitive performances.


\end{abstract}

\section{Introduction}

The EigenDecomposition (ED) or the Singular Value Decomposition (SVD) explicitly factorize a matrix into the eigenvalue and eigenvector matrix, which serves as a fundamental tool in computer vision and deep learning. Recently, many algorithms integrated the SVD as a meta-layer into their models to perform some desired spectral transformations~\cite{lin2015bilinear,lin2017improved,li2017second,brachmann2017dsac,huang2018decorrelated,cho2019image,wang2019backpropagation,huang2020investigation,campbell2020solving,dang2020eigendecomposition,wang2020diversified,wang2021robust,song2021approximate}. The applications vary in global covariance pooling~\cite{li2017second,wang2020deep,song2021approximate}, decorrelated Batch Normalization (BN)~\cite{huang2018decorrelated,wang2019backpropagation,huang2020investigation,song2022fast}, Perspective-n-Points (PnP) problems~\cite{brachmann2017dsac,campbell2020solving,dang2020eigendecomposition}, and Whitening and Coloring Transform (WCT)~\cite{li2017universal,cho2019image,wang2020diversified}. 

\begin{figure}[htbp]
    \centering
    \includegraphics[width=0.8\linewidth]{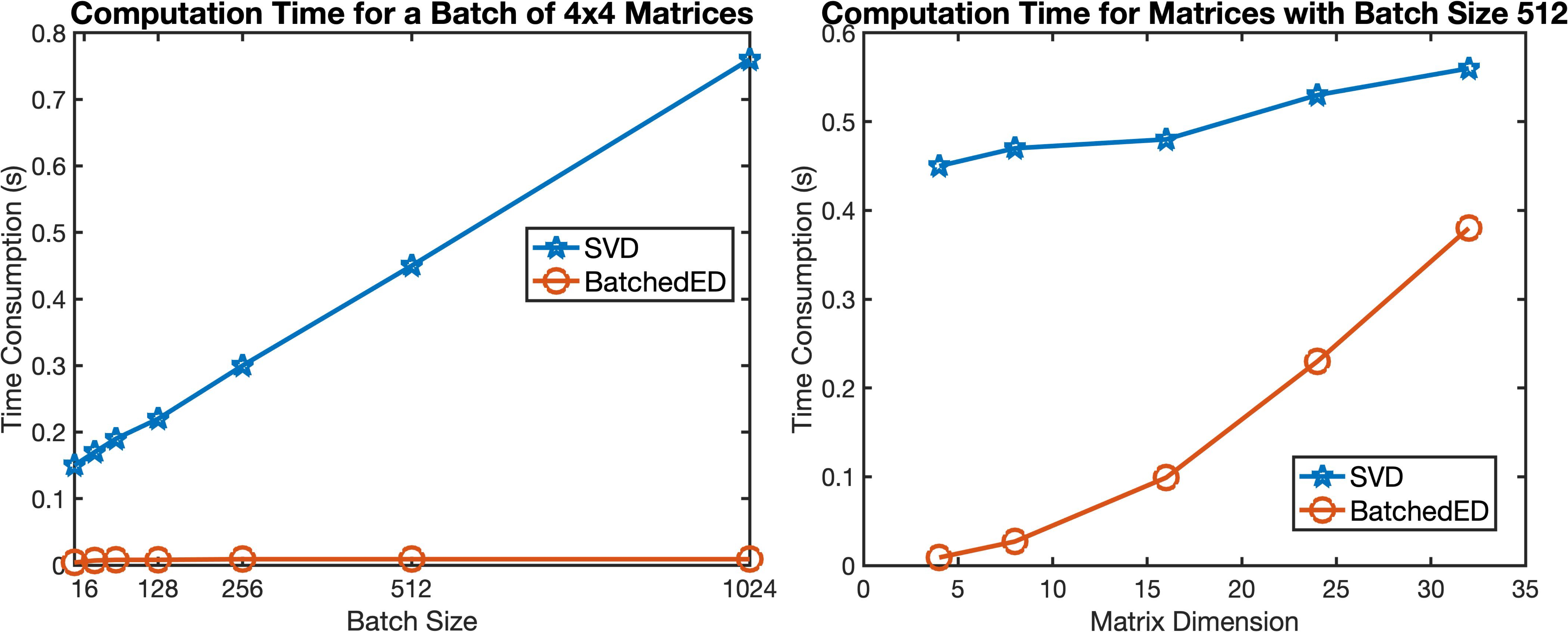}
    \caption{The speed comparison of our Batched ED against the \textsc{torch.svd}. (\emph{Left}) Time consumption for a mini-batch of $4{\times}4$ matrices with different batch sizes. (\emph{Right}) Time consumption for matrices with batch size $512$ but in different matrix dimensions.}
    \label{fig:cover}
    \vspace{-0.4cm}
\end{figure}

The problem setup of the ED in computer vision is quite different from other fields. In other communities such as scientific computing, batched matrices rarely arise and the ED is usually used to process a single matrix. However, in deep learning and computer vision, the model takes a mini-batch of matrices as the input, which raises the requirement for an ED solver that works for batched matrices efficiently. Moreover, the differentiable ED works as a building block and needs to process batched matrices millions of times during the training and inference. This poses a great challenge to the efficiency of the ED solver and could even stop people from adding the ED meta-layer in their models due to the huge time consumption (see Fig.~\ref{fig:cover}).

In the current deep learning frameworks such as Pytorch~\cite{paszke2019pytorch} or Tensorflow~\cite{abadi2016tensorflow}, the ED solvers mainly adopt the SVD implementation from the linear algebra libraries (\emph{e.g.}, LAPACK~\cite{anderson1999lapack} and Intel MKL~\cite{wang2014intel}). These solvers can efficiently process a single matrix but do not support batched matrices on GPUs well. Most of the implementations are based on the Divide-and-Conquer (DC) algorithm~\cite{cuppen1980divide,gu1995divide}. This algorithm partitions a matrix into multiple small sub-matrices and performs the ED simultaneously for each sub-matrix. Aided by the power of parallel and distributed computing, its speed is only mildly influenced by the matrix dimension and can be very fast for a single matrix. The core of the DC algorithm is the characteristic polynomials $\det(\lambda\mathbf{I}{-}\mathbf{A}){=}0$, which can be solved by various methods, such as secular equations~\cite{gu1995divide} and spectral division~\cite{nakatsukasa2013stable}. However, solving the polynomial requires simultaneously localizing all the eigenvalue intervals for each individual matrix. Despite the high efficiency for a single matrix, these DC algorithms do not scale to batched matrices. 

Except for the DC algorithm, some ED solvers would use the QR iteration. The QR iteration has many implementation methods and one particular batch-efficient choice is by Givens rotation. The Givens rotation can be implemented via matrix-matrix multiplications, which naturally extends to batched matrices. During the QR iterations, the Givens rotation is applied successively to annihilate the off-diagonal entries until the matrix becomes diagonal. The major drawback limiting the usage of QR iterations is the $O{(}n^5{)}$ time cost, which makes this method only applicable to tiny matrices (\emph{e.g.,} $dim{<}9$). To alleviate this issue, modern QR-based ED implementations apply the technique of \emph{deflation}~\cite{ahues1997new,braman2002multishift,braman2002multishift2}, \emph{i.e.,} partition the matrix into many sub-matrices. The deflation technique can greatly improve the speed of the QR iterations but only works for an individual matrix. For the QR iteration, the convergence speed is related with the adjacent eigenvalue ratio $\frac{\lambda_{i{+}1}}{\lambda_{i}}$. For multiple matrices within a mini-batch, the off-diagonal entries of each matrix converge to zero with inconsistent speed and where each matrix can be partitioned is different. Consequently, the deflation technique does not apply to batched matrices either. To give a concrete example, consider $2$ matrices of sizes $8{\times}8$ in a mini-batch. Suppose that the deflation would split one into two $3{\times}3$ and $5{\times}5$ matrices, while the other matrix might be partitioned into two $4{\times}4$ matrices. In this case, the partitioned matrices cannot be efficiently processed as a mini-batch due to the inconsistent matrix sizes.

To attain a batch-friendly and GPU-efficient ED method dedicated to computer vision field, we propose a QR-based ED algorithm that performs the ED via batched matrix/vector multiplication. Each step of the ED algorithm is carefully motivated for the best batch-efficient and computation-cheap consideration. We first perform a series of batched Householder reflectors to tri-diagonalize the matrix by the batched matrix-vector multiplication. Afterward, the explicit QR iteration by matrix rotation with double Wilkinson shifts~\cite{wilkinson1971algebraic} is conducted to diagonalize the matrix. The proposed shifts make the last two diagonal entries of the batched matrices have consistent convergence speed. Thereby the convergence is accelerated and the matrix dimension can be progressively shrunk during the QR iterations. Besides the dimension reduction, we also propose some economic computation methods based on the complexity analysis. The time complexity of QR is thus reduced from $O{(}n^5{)}$ to $O{(}n^3{)}$. The numerical tests demonstrate that, for matrices whose dimensions are smaller than $24$, our Pytorch implementation is consistently much faster than the default SVD routine for any batch size. For matrices with larger dimensions (\emph{e.g.,} $dim{=}32$ or $36$), our method could also have an advantage when the batch size is accordingly large (see also Fig.~\ref{fig:cover}). We validate the effectiveness of our method in several applications of differentiable SVD, including decorrelated BN, covariance pooling for vision transformers, and neural style transfer. Our Batched ED achieves competitive performances against the SVD.

The contributions of the paper are summarized threefold:
\begin{itemize}
    \item We propose an ED algorithm for a mini-batch of small and medium matrices which is dedicated to many application scenarios of computer vision. Each step of ED is carefully motivated and designed for the best batch efficiency.
    \item We propose dedicated acceleration techniques for our Batched ED algorithm. The progressive dimension shrinkage is proposed to reduce the matrix size during the iterations, while some economic computation methods grounded on the complexity analysis are also developed.
    \item Our batch-efficient ED algorithm is validated in several applications of differentiable SVD. The experiments on visual recognition and image generation demonstrate that our method achieves very competitive performances against the SVD encapsuled in the current deep learning platforms.
\end{itemize}

\section{Related Work}
In this section, we discuss the related work in computing the differentiable ED and its applications.

\subsection{Computing the Differentiable ED}

To perform the ED, modern deep learning frameworks (\emph{e.g.,} Pytorch and Tensorflow) call the LAPACK's SVD routine by default. The routine uses the Divide-and-Conquer algorithm~\cite{cuppen1980divide,gu1995divide} to conduct the ED.  Assisted by the power of parallel and distributed computing, the divide-and-conquer-based ED can simultaneously process each sub-matrix and achieve high efficiency for a single matrix regardless of the matrix size. However, solving the core characteristic polynomials requires simultaneously finding all the eigenvalue intervals for each individual matrix, which causes this algorithm unable to scale to batched matrices well. There are also some routines that use QR iterations with deflation for performing the ED~\cite{braman2002multishift,braman2002multishift2}. Equipped with the deflation technique to partition the matrices, the QR iteration can also have a fast calculation speed. When it comes to a mini-batch of matrices, the off-diagonal entries of each matrix converge to zero with different speeds, and where each matrix can be partitioned is inconsistent. Hence, the deflation technique cannot be applied to batched matrices.

For the back-propagation of the ED, it suffers from the numerical instability caused by the close and repeated eigenvalues. Recently, several methods have been proposed to solve the instability issue~\cite{wang2019backpropagation,wang2021robust,song2021approximate}. Wei~\emph{et al.}~\cite{wang2019backpropagation} propose to use Power Iteration (PI) to approximate the SVD gradients. Song~\emph{et al.}~\cite{song2021approximate} propose to use Pad\'e approximants to closely estimate the gradients. Despite the applicability of these methods, a more practical approach is to divide the features $\mathbf{X}{\in}{\mathrm{R}^{C{\times}BHW}}$ into groups $\mathbf{X}{\in}{\mathrm{R}^{G{\times}{\frac{C}{G}}{\times}BHW}}$ in the channel dimension and attain a mini-batch of small covariance matrices $\mathbf{X}\mathbf{X}^{T}{\in}{\mathrm{R}^{G{\times}{\frac{C}{G}}{\times}{\frac{C}{G}}}}$~\cite{huang2018decorrelated,pan2019switchable}, which can keep more channel statistics and naturally avoid the gradient explosion issue caused by the rank-deficiency. This also raises the need of such an ED solver that works for batched matrices efficiently. 

To attain the batch-efficient ED algorithm dedicated to computer vision field, we propose our QR-based algorithm for small and medium batched matrices. We motivate each step of our ED algorithm for the best batch efficiency. Our ED solver integrates double Wilkinson shifts~\cite{wilkinson1971algebraic} to guarantee that the last two diagonal entries have consistent convergence speed within the mini-batch, and consequently the matrix dimension can be progressively reduced. With several other acceleration technique grounded on the complexity analysis, our solver can be much faster than Pytorch SVD for a mini-batch of small matrices.

\subsection{Applications of the Differentiable ED}

The need for differentiable ED arises in numerous applications of computer vision. Some methods adopt the end-to-end ED to compute the matrix square root of the global covariance feature before the fully-connected layer~\cite{lin2015bilinear,li2017second,li2018towards,wang2020deep,song2021approximate,xie2021so,gao2021temporal,song2022fast2,song2022eigenvalues}. Such approaches are termed as Global Covariance Pooling (GCP) methods, and they have achieved state-of-the-art performances on both generic and fine-grained visual recognition. Another line of research uses the ED to perform the decorrelated batch normalization (BN)~\cite{huang2018decorrelated,huang2019iterative,pan2019switchable,huang2020investigation,song2022fast,huang2021group,song2022fast}. The process resembles the ZCA whitening transform to compute the inverse square root for eliminating the correlation between features. The differentiable ED can be also applied in the area of neural style transfer. As pointed out in~\cite{gatys2015texture,gatys2016image}, the feature covariance naturally embeds the style information. Some methods use the differentiable ED to perform successive WCT on the feature covariance for the universal style transfer~\cite{li2017universal,cho2019image,choi2021robustnet}. In the geometric vision, the ED is often applied to solve the PnP problem and estimate the camera pose~\cite{lepetit2009epnp,brachmann2017dsac,campbell2020solving,dang2020eigendecomposition}. Besides the main usages above, there are some other minor applications~\cite{Dai_2019_CVPR,sun2021second}. 

\section{Methodology}

\begin{figure*}[t]
    \centering
    \includegraphics[width=0.99\linewidth]{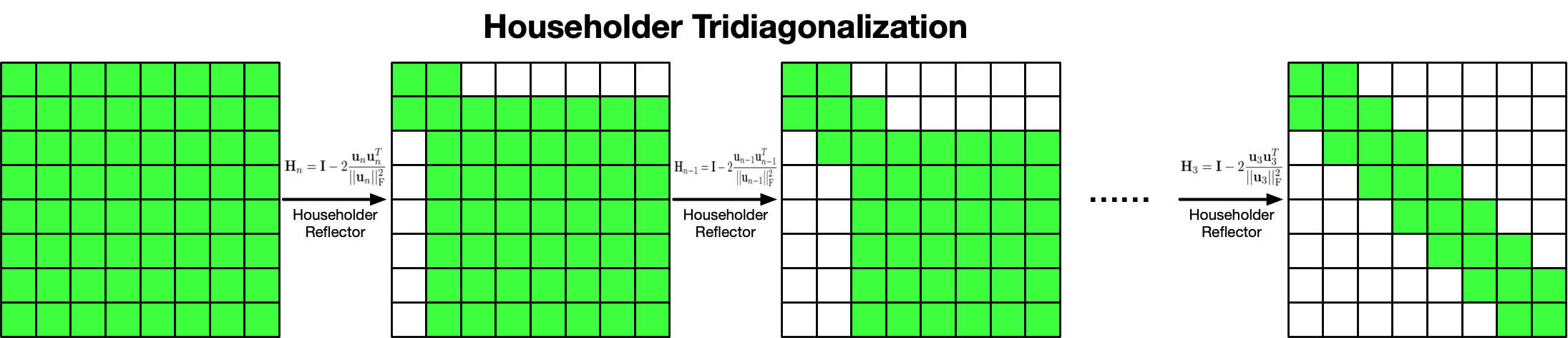}
    \caption{Visual illustration of our batched Householder tri-diagonalization. After ${(}n{-}2{)}$ designed reflections, the symmetric matrix $\mathbf{A}$ is reduced to a tri-diagonal matrix $\mathbf{T}$.}
    \label{fig:householder}
\end{figure*}

In this section, we present our method that performs the Batched ED. Our algorithm is implemented via the sequential batched Householder reflectors to tri-diagonalize the matrix and the batched QR iteration to diagonalize the tri-diagonal matrix. Both processes are GPU-friendly and batch-efficient. Now we illustrate each process in detail. Notice that every step is applied on batched matrices for the best efficiency.

\subsection{Batched Tri-diagonalization based on Householder Reflection} 
Given the Hermitian matrix $\mathbf{A}$, the tri-diagonalization process is defined as:
\begin{equation}
    \mathbf{A}=\mathbf{P}\mathbf{T}\mathbf{P}^{T}
\end{equation}
where $\mathbf{T}$ is a tri-diagonal matrix, and $\mathbf{P}$ is an orthogonal matrix. To perform such an orthogonal similarity transform, we can decompose $\mathbf{P}$ into $n{-}2$ Householder reflectors. This leads to the re-formulation:
\begin{equation}
    \mathbf{T}=\mathbf{P}^{T}\mathbf{A}\mathbf{P}=(\mathbf{H}_{n}\dots\mathbf{H}_{4}\mathbf{H}_{3})^{T}\mathbf{A}(\mathbf{H}_{n}\dots\mathbf{H}_{4}\mathbf{H}_{3})
\end{equation}
Each reflector is both orthogonal ($\mathbf{H}\mathbf{H}^{T}{=}\mathbf{I}$) and unitary ($\mathbf{H}{=}\mathbf{H}^{T}$). The reflector is constructed using an vector:
\begin{equation}
    \mathbf{H}=\mathbf{I}-2\frac{\mathbf{u}\mathbf{u}^{T}}{||\mathbf{u}||_{\rm F}^{2}}
\end{equation}
The matrix $\mathbf{H}$ reflects the vector $\mathbf{u}$ along the direction that is perpendicular to the hyper-plane orthogonal to $\mathbf{u}$. This property can be used to tri-diagonalize a symmetric matrix by reflecting each row and column sequentially. A Householder reflection is computed by:
\begin{equation}
    \begin{aligned}
    \mathbf{H}\mathbf{A}\mathbf{H}&=(\mathbf{I}-2\frac{\mathbf{u}\mathbf{u}^{T}}{||\mathbf{u}||_{\rm F}^{2}})\mathbf{A}(\mathbf{I}-2\frac{\mathbf{u}\mathbf{u}^{T}}{||\mathbf{u}||_{\rm F}^{2}})\\
    &=\mathbf{A}-\mathbf{p}\mathbf{u}^{T}-\mathbf{u}\mathbf{q}^{T}
    \label{rank_2_update}
    \end{aligned}
\end{equation}
where the temporary variables $\mathbf{q}$, $\mathbf{p}$, and $K$ are defined as:
\begin{equation}
    \mathbf{q}=\mathbf{p} - K\mathbf{u}, \mathbf{p}=\frac{2\mathbf{A}\mathbf{u}}{||\mathbf{u}||_{\rm F}^{2}}, K=\frac{\mathbf{u}^{T}\mathbf{p}}{||\mathbf{u}||_{\rm F}^{2}}
    \label{temp_variable}
\end{equation}
As can be seen, \cref{rank_2_update} actually defines a symmetric rank-2 update on $\mathbf{A}$. By some deductions on~\cref{rank_2_update}, each Householder reflector can be designed to introduce zero entries to a row and a column (see Fig.~\ref{fig:householder}). We omit the derivation of the vector for conciseness and give the result here:
\begin{equation}
    \begin{gathered}
    \mathbf{u}_{i}=[0,\dots,a_{i,i}, a_{i,i+1},\dots ,a_{i,n-1}, \sigma],\\
    \sigma =\pm \sqrt{a_{i,i}^2 + a_{i,i+1}^2 + \dots + a_{i,n-1}^2}
    \end{gathered}
\end{equation}
where $a_{i,j}$ denotes the entry of $\mathbf{A}$ at $i$-th row and $j$-th column, and the sign of $\sigma$ is usually chosen as ${\rm sign}(a_{i,n})$ to reduce the round-off error. 
By such a construction, only $n{-}2$ reflections are needed to transform the symmetric matrix into the tri-diagonal form. Each householder reflection needs $2$ matrix-matrix multiplication, which takes $O{(}2n^{3}{)}$ complexity. However, as indicated in~\cref{rank_2_update} and \cref{temp_variable}, the calculation can be reduced to one matrix-matrix multiplication and two matrix-vector multiplication, which needs the complexity of $O{(}n^{3}{+}2n^{2}{)}$. 

When the eigenvector is required, we can calculate $\mathbf{P}$ by accumulating the Householder reflectors:
\begin{equation}
    \mathbf{P} = \mathbf{H}_{n}\dots\mathbf{H}_{4}\mathbf{H}_{3}
\end{equation}
The computation needs ${(}n{-}2{)}$ matrix multiplication where the complexity of each multiplication is $O{(}n^{3}{)}$. We note this step can be further accelerated by: 

\begin{thm}[WY representation~\cite{bischof1987wy}]
 For any accumulation of $m$ Householder matrices $\mathbf{H}_{1}{\dots}\mathbf{H}_{m}$, there exists $\mathbf{W}{,}\mathbf{Y}{\in}\mathrm{R}^{{(}n{-}2{)}{\times}m}$ such that we have the relation $\mathbf{I}{-}2\mathbf{W}\mathbf{Y}^{T}{=}\mathbf{H}_{1}{\dots}\mathbf{H}_{m}$. Computing $\mathbf{W}$ and $\mathbf{Y}$ takes $O{(}{(}n{-}2{)}m{)}$ time and ${(}m{-}1{)}$ sequential Householder multiplications.
\end{thm}

Relying on this theorem, we can divide the accumulation $\mathbf{H}_{n}{\dots}\mathbf{H}_{4}\mathbf{H}_{3}$ into $\nicefrac{(n-2)}{m}$ sub-sequences and compute them in parallel. Each sub-sequence takes $O{(}{(}m{-}1{)}n^{3}{+}{(}n{-}2{)}m{)}$ time to compute the WY representation and $O{(}{(}n{-}2{)^{2}m}{)}$ time to compute $\mathbf{I}{-}2\mathbf{W}\mathbf{Y}^{T}$. Combining all the sub-sequence needs extra time of $O{(}\nicefrac{(n{-}2)^3}{m}{)}$.
This can further reduce the complexity of computing $\mathbf{P}$ from $O{(}{(}n{-}3{)}n^{3}{)}$ to $O{(}{(}m{-}1{)}n^{3}{+}{(}n{-}2{)}m{+}{(}n{-}2{)^{2}m}{+}\nicefrac{(n{-}2)^3}{m}{)}$. The computation saving would be huge when $n$ is large.

\subsection{Batched Diagonalization based on QR Iteration}

After obtaining the tri-diagonal matrix $\mathbf{T}$, we use the Givens rotation to perform the QR iterations, which can be implemented efficiently via batched matrix multiplication. Based on the ordinary QR iteration, we further apply several techniques to speed up the convergence and save the computational budget.

\subsubsection{Basic QR Iteration by Givens Rotation.} Given the tri-diagonal matrix $\mathbf{T}$, the QR iteration takes the following iterative update:
\begin{equation}
    \begin{gathered}
    \mathbf{T}_{k}=\mathbf{Q}_{k}\mathbf{R}_{k},\ \mathbf{T}_{k+1}=\mathbf{R}_{k}\mathbf{Q}_{k} 
    \end{gathered}
    \label{qr_2step}
\end{equation}
where $\mathbf{Q}_{k}$ denotes the orthogonal matrix, and $\mathbf{R}_{k}$ is the upper-triangular matrix. Replacing $\mathbf{R}_{k}$ with $\mathbf{Q}_{k}^{T}\mathbf{T}_{k}$ leads to the re-formulation of~\cref{qr_2step} :
\begin{equation}
    \mathbf{T}_{k+1}=\mathbf{Q}_{k}^{T}\mathbf{T}_{k}\mathbf{Q}_{k}
\end{equation}
As can be seen, a single QR iteration is equivalent to performing an orthogonal similarity transform. By performing the iterations, the sub-diagonal and super-diagonal entries are gradually reduced till the matrix becomes diagonal. 

\begin{figure*}[t]
    \centering
    \includegraphics[width=0.9\linewidth]{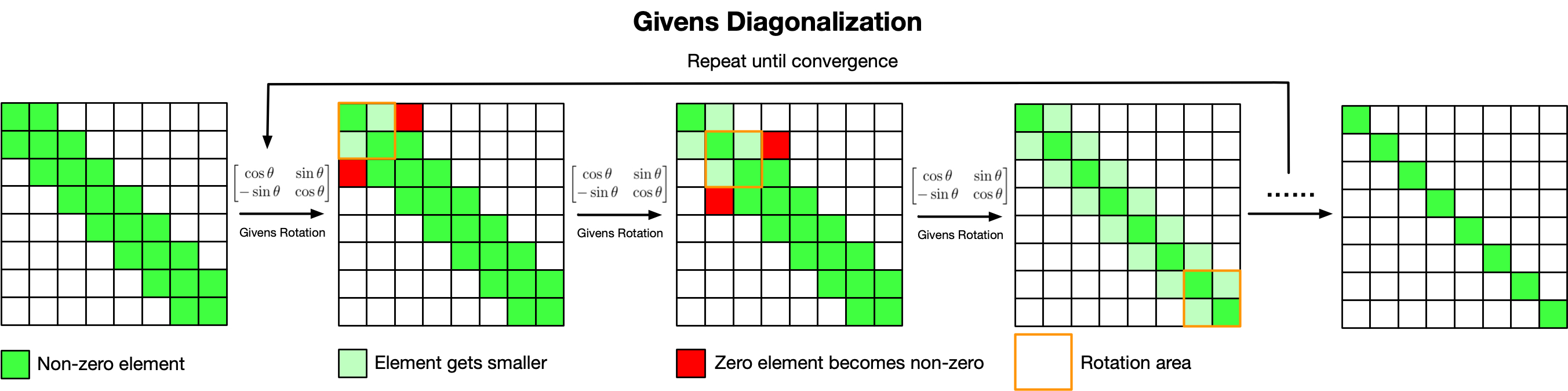}
    \caption{Visual illustration of the batched Givens diagonalization. For each QR iteration, the Givens rotation is applied from the left top corner to the bottom right corner to reduce the magnitude of the off-diagonal elements. The iteration continues till the off-diagonal entries become zero or below a certain tolerance.}
    \label{fig:givens}
\end{figure*}

For each QR iteration, we construct the orthogonal transform using successive Givens rotations moving from the left top corner to the right bottom corner. The $2{\times}2$ Givens Rotation and its $n{\times}n$ extension are defined by:
\begin{equation}
     \mathbf{R}_{2\times2}=\begin{bmatrix}
    \cos{\theta} & -\sin{\theta} \\
    \sin{\theta} & \cos{\theta} \\
    \end{bmatrix}, \mathbf{R}_{n\times n}=\begin{bmatrix}
    \mathbf{I} & \mathbf{0} & \mathbf{0} \\
    \mathbf{0} & \mathbf{R}_{2\times 2} & \mathbf{0} \\
    \mathbf{0} & \mathbf{0} & \mathbf{I}\\
    \end{bmatrix}
\end{equation}
where $\theta$ is the rotation angle, and the rotation matrix is orthogonal but not symmetric (\emph{i.e.,} $\mathbf{R}^{T}\mathbf{R}{=}\mathbf{I}$ and $\mathbf{R}^{T}{\neq}\mathbf{R}$). As shown in Fig.~\ref{fig:givens}, by design of the rotation angle, the successive Givens Rotation applied on $\mathbf{T}$ can keep the tri-diagonal form but reduce the magnitude of off-diagonal elements. For the derivation of Givens rotation, please refer to the supplementary material for detail. The sequential Givens rotations moving along the diagonal form one single QR iteration:
\begin{equation}
    \mathbf{T}_{k+1} = (\mathbf{R}_{n-2}^{T}\dots\mathbf{R}_{0}^{T})\mathbf{T}_{k}(\mathbf{R}_{0}\dots\mathbf{R}_{n-2}) 
\end{equation}
where $\mathbf{R}_{i}$ denotes the $i$-th rotation counting from the left top corner. For the orthogonal matrix $\mathbf{Q}_{i}$ in the $i$-th QR iteration, we can easily find out
\begin{equation}
    \mathbf{Q}_{k}=\mathbf{R}_{0}\dots\mathbf{R}_{n-2}
\end{equation}
Taking the Householder tri-diagonalization and Givens diagonalization together, our batch-efficient ED algorithm can be formally defined by:
\begin{equation}
    \mathbf{A}=(\mathbf{P}\mathbf{Q}_{0}\dots\mathbf{Q}_{k}) \mathbf{\Lambda} (\mathbf{P}\mathbf{Q}_{0}\dots\mathbf{Q}_{k})^{T}
\end{equation}
where $k$ is the iteration times of the QR iteration, $\mathbf{\Lambda}$ is the eigenvalue matrix, and $\mathbf{P}\mathbf{Q}_{0}{\dots}\mathbf{Q}_{k}$ is the eigenvector matrix. For the convergence, we have:

\begin{thm}[Convergence of QR iteration]
\label{QR_convergence}
Let $\mathbf{T}$ be the positive definite tri-diagonal matrix with the eigendecomposition $\mathbf{Q}\mathbf{\Lambda}\mathbf{Q}^{T}$ and assume $\mathbf{Q}^{T}$ can be LU decomposed. Then the QR iteration of $\mathbf{T}$ will converge to $\mathbf{\Lambda}$.
\end{thm}

We defer the proof to the Supplementary Material. The key results of this theorem is that the convergence speed depends on the adjacent eigenvalue ratio $\frac{\lambda_{i}}{\lambda_{j}}$ for $i{>}j$. The QR iterations usually take $2n$ iterations to make the resultant matrix diagonal~\cite{francis1962qr}. Consider the fact that each iteration takes ${(}n{-}1{)}$ Givens rotation. The computation overhead would be huge. For deriving the eigenvalues, we need $4n{(}n{-}1{)n^3}$ time, while it takes the complexity of $2n{(}n{-}2{)n^3}{+}{(}2n{-}1{)}n^3$ to compute the eigenvector. The time complexity of the QR iteration is quintic to the matrix dimension $n$, which would make this method only applicable to the tiny matrices (${<}9$). Existing deflation techniques~\cite{braman2002multishift,braman2002multishift2} to accelerate the computation cannot be applied to our batched matrices. To resolve this issue, we propose the following techniques:


\subsubsection{Double Wilkinson Shift.} As indicated in Theorem~\ref{QR_convergence}, the convergence speed of QR iteration depends on the ratio $\frac{\lambda_{i}}{\lambda_{j}}$, where $i{>}j$. A natural approach to accelerate the convergence is to shift the matrix by $\mathbf{T}-\mu\mathbf{I}$ such that the convergence speed becomes $\frac{\lambda_{i}-\mu}{\lambda_{j}-\mu}$. A preferable shift coefficient should be $u{=}\lambda_{j}$, as this can help the matrices to converge quickly: $\frac{\lambda_{i}-\mu}{\lambda_{j}-\mu}{=}\infty$. This is particularly useful for matrices in a mini-batch as the speed can be made consistent by shifting. 

Since each Givens rotation will affect the area rotated by the previous one, only the last $2{\times}2$ Givens rotation will not be influenced, \emph{i.e.,} the two eigenvalues of the last block can be locally estimated. Thus, we propose to extract the shift coefficients from the $2{\times}2$ block on the right bottom corner:
\begin{equation}
    \mu_{n-2}, \mu_{n-1}=Wilkinson(\mathbf{T}_{k}[n-2:n])
\end{equation}
where $\mathbf{T}_{k}{[}n{-}2{:}n{]}$ denotes the last $2{\times}2$ block of $\mathbf{T}_{k}$, and $\mu_{n-2}$ and $\mu_{n-1}$ are the two eigenvalues computed from this block. These shifting coefficients are referred to as the Wilkinson shift~\cite{wilkinson1971algebraic}, and we give the derivation in the supplementary material. After attaining the shifts, we can reformulate the QR iterations with double shifts:
\begin{equation}
    \begin{gathered}
     \mathbf{T}_{k+\nicefrac{1}{2}}=\mathbf{Q}_{k}^{T}(\mathbf{T}_{k}-\mu_{n-1}\mathbf{I})\mathbf{Q}_{k}+\mu_{n-1}\mathbf{I}\\
     \mathbf{T}_{k+1}= \mathbf{Q}_{k}^{T}(\mathbf{T}_{k+\nicefrac{1}{2}}-\mu_{n-2}\mathbf{I})\mathbf{Q}_{k}+\mu_{n-2}\mathbf{I}
    \end{gathered}
\end{equation}
With the shifts, the integrated iteration consists of two sequential QR iterations shifted by the eigenvalues $\mu_{n-2}$ and $\mu_{n-1}$, respectively. 

\subsubsection{Progressive Dimension Shrinkage.}

\begin{wrapfigure}{r}{0.4\textwidth}
    \centering
    \includegraphics[width=0.99\linewidth]{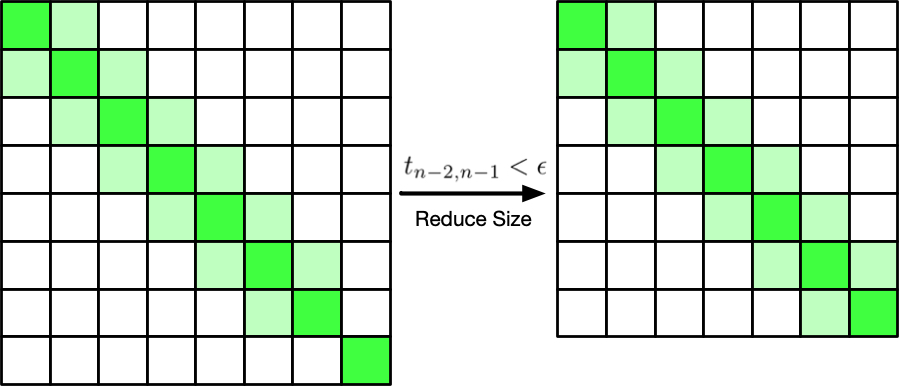}
    \caption{Illustration of the progressive dimension reduction in the QR iterations. After one iteration, if the last sub-diagonal entry is below a small threshold $\epsilon$, we can remove the last row and column.}
    \label{fig:dim_reduction}
\end{wrapfigure}

One direct benefit brought by the Wilkinson shift is that, for all the matrices in a mini-batch, the last two diagonal entries can quickly converge to the corresponding eigenvalues and the off-diagonal elements can converge to zero:
\begin{equation}
    \begin{gathered}
 \!\! t_{n{-}2,n{-}2}{\rightarrow}\lambda_{n-2},\  t_{n{-}2,n{-}3}{=}t_{n{-}3,n{-}2}{\rightarrow}0\\
   \!\! t_{n{-}1,n{-}1}{\rightarrow}\lambda_{n-1},\ t_{n{-}2,n{-}1}{=}t_{n{-}1,n{-}2}{\rightarrow}0
    \end{gathered}
\end{equation}
We can use this property to speed up the computation by gradually reducing the matrix dimension, \emph{i.e.,} shrinking the matrix by $\mathbf{T}{\in}\mathrm{R}^{n{\times n}}{\rightarrow}\mathbf{T}{\in}\mathrm{R}^{{(}n{-}1{)}{\times}{(}n{-}1{)}}$ after one iteration. As shown in Fig.~\ref{fig:dim_reduction}, when the last sub-diagonal entry is below a given small threshold (\emph{e.g.,} $1e{-}5$), we could shrink the matrix by removing the last row and column. In doing so, the matrix size is progressively reduced during the QR iterations. With the dimension reduction, one QR iteration would take ${(}n{-}1{-}r{)}$ Givens rotations, where $r$ is the reduction times.

\subsubsection{Economic Eigenvalue Calculation.} For a Givens rotation, it only affects the adjacent $4{\times}4$ block. We can save the computation budget by applying the matrix multiplication on the $4{\times}4$ rotation region in the neighborhood. This reduces the time of a rotation from $O{(}2n^3{)}$ to $O{(}2{\times}4^3{)}{=}O{(}128{)}$, which makes each rotation consume a constant time cost. Taking the above dimension reduction into account, the QR iterations need $O{(}256n{(}n{-}1{-}r{)}{)}$ time to derive the eigenvalues.

\subsubsection{Economic Eigenvector Calculation.} Equipped with the progressive dimension reduction, the orthogonal transform $\mathbf{Q}_{k}$ in a QR iteration is defined by:
\begin{equation}
    \mathbf{Q}_{k}=\mathbf{R}_{0} \mathbf{R}_{1} \dots \mathbf{R}_{n-2-r}
    \label{q_eigenvector}
\end{equation}
where we need ${(}n{-}2{-}r{)}$ rotations for each iteration. The computation can be potentially simplified by the theorem:

\begin{thm}[Implicit Q Theorem~\cite{golub1996matrix}]
Let $\mathbf{B}$ be an upper Hessenberg and only have positive elements on its first sub-diagonal. Assume there exists a unitary transform $\mathbf{Q}^{H}\mathbf{A}\mathbf{Q}{=}\mathbf{B}$. Then $\mathbf{Q}$ and $\mathbf{B}$ are uniquely determined by $\mathbf{A}$ and the first column of $\mathbf{Q}$.
\end{thm}

We give the proof and some discussion in the supplementary document. This theorem implies that, without the need of explicit QR iteration, the orthogonal transform $\mathbf{Q}$ and the transformed matrix $\mathbf{B}$ can be both implicit calculated. However, it assumes that the sub-diagonal elements of $\mathbf{B}$ are positive. In our case, the Givens rotation can easily zero out the last two sub-diagonal entries. Consequently, directly using the theorem would cause large round-off errors and data overflow.

Although the theorem cannot be directly applied, it allows us to simplify the eigenvector calculation. As indicated by the theorem, the $i$-th rotation would only affect the orthogonal matrix $\mathbf{Q}$ on the area after the $i$-th row and column. We can reduce the computation by involving only part of the matrix and simplify the calculation in~\cref{q_eigenvector} as:
\begin{equation}
    \mathbf{Q}_{k}=\mathbf{R}_{0}{[}1{:}{]} \mathbf{R}_{1}{[}2{:}{]} \dots \mathbf{R}_{n-2-r}[n{-}2{-}r{:}]
\end{equation}
where ${[}i{:}{]}$ denotes part of the matrix that excludes the first $i$ rows and columns. By doing so, the time complexity of calculating $\mathbf{Q}_{k}$ can be reduced to:
\begin{equation}
    \begin{gathered}
    (n{-}2{-}r)^2 n {+} (n{-}3{-}r)^2 n {+} \dots {+} 1^2 n\\
    {=}{\sum_{i=1}^{n{-}2{-}r}}{i^2 n}{=}\frac{(n{-}2{-}r)(n{-}1{-}r)(2n{-}3{-}2r)}{6}n
    \end{gathered}
\end{equation}
Compared with the original time cost $O{(}(n{-}2{-}r)n^3{)}$, the saving would be considerable for large $n$ and $r$.

\subsection{Computation Complexity Summary}

\begin{table*}[htbp]
    \centering
    \caption{Comparison of time complexity of the basic QR-based ED solver and our ED solver dedicated for batched matrices. Here $n$ denotes the matrix size and $r$ represents the average reduction times during the QR iterations.}
    \label{tab:computation_complexity}
    \resizebox{0.9\linewidth}{!}{
    \begin{tabular}{r|c|c}
    \toprule
       \multirow{2}*{Time} &  \multicolumn{2}{c}{Basic QR-based ED Solver}  \\
       \cline{2-3}
        & Eigenvalue & Eigenvector\\
        \hline
        Tri-diag. &$2n^{3}$ &${(}n{-}3{)}n^{3}$\\
        QR &$4n{(}n{-}1{)n^3}$ &$2n{(}n{-}2{)n^3}{+}{(}2n{-}1{)}n^3$ \\
        \hline
        Sum & ${(}4n^2{-}4n{+}2{)n^3}$ &${(}2n^2{-}n{-}4{)n^3}$ \\
        \hline
        \hline
        \multirow{2}*{Time} & \multicolumn{2}{c}{Our Batched ED Solver}\\
        \cline{2-3}
        &Eigenvalue & Eigenvector \\
        \hline
        Tri-diag.  &$n^{3}{+}2n^{2}$ &${(}m{-}1{)}n^{3}{+}{(}n{-}2{)}m{+}{(}n{-}2{)^{2}m}{+}\frac{(n{-}2)^3}{m}$ \\
        QR &$256{(}n{-}1{-}r{)}n$ &$\frac{(n{-}2{-}r)(n{-}1{-}r)(2n{-}3{-}2r)}{6}2n^2{+}{(}2n{-}1{)}n^3$ \\
        \hline
        Sum & $n^{3}{+}258n^{2}{-}256n{(}1{+}r{)}$ & \begin{tabular}{c}$\frac{2}{3}n^5{-}{(}2r{+}1{)}n^4{+}{(}2r^2{+}6r{+}\frac{7}{3}{+}m{)}n^3$ \\ ${-}{(}\frac{3}{2}r^3{+}3r^2{+}\frac{13}{3}r{+}2{-}m{)}n^2{-}{(}3m{-}\frac{3}{m}{)}n{+}2m{-}\frac{6}{m}$ \end{tabular}\\
        \bottomrule
    \end{tabular}
    }
\end{table*}

Table~\ref{tab:computation_complexity} summarizes the time complexity of the basic QR-based ED solver and our proposed ED solver dedicated for batched matrices. Taking the highest-order term for simpler analysis, our ED solver reduces the time from $O{(}4n^5{)}$ to $O{(}n^3{)}$ for computing the eigenvalues, and saves the time from $O{(}2n^5{)}$ to $O{(}\frac{2}{3}n^5{)}$ for eigenvectors. Moreover, depending on the reduction times $r$, the complexity can be further reduced with the term ${-}256{(}1{+}r{)}n$ for eigenvalues and ${-}{(}2r{+}1{)}n^4$ for eigenvectors.


\subsection{Convergence and Error Bounds}
For the tri-diagonalization process, the convergence is guaranteed with $n{-}2$ Householder reflectors. The error is only related to the machine precision and data precision, which can be sufficiently neglected. For the QR iterations, the convergence mainly depends on the adjacent eigenvalue ratio $\frac{\lambda_{i+1}}{\lambda_{i}}$ and the shift $\mu$. In certain cases when the two eigenvalue are close ($\frac{\lambda_{i+1}}{\lambda_{i}}{\approx}1$), the convergence speed is slow and the residual term $(\frac{\lambda_{i+1}{-}\mu}{\lambda_{i}-\mu})^{2n}$ becomes the error. Another error source comes from the tolerance $\epsilon$ for the dimension reduction. Let $\Bar{\mathbf{\Lambda}}$ represent the exact eigenvalues and ${\mathbf{\Lambda}}$ denote the eigenvalues calculated by our ED solver. Then the error is bounded by:
\begin{equation}
    ||\Bar{\mathbf{\Lambda}}-\mathbf{\Lambda}||_{\rm F}\leq\max_{i}((\frac{\lambda_{i+1}-\mu}{\lambda_{i}-\mu})^{2n} |l_{i+1,i}| )+\epsilon
\end{equation}
where $l_{i{+}1{,}i}$ is the entry of $\mathbf{L}$ computed by $\mathbf{Q}^{T}{=}\mathbf{L}\mathbf{U}$, and the shift $\mu$ changes every QR iteration. Since $\mathbf{Q}$ is orthogonal, the magnitude of $l_{i{+}1{,}i}$ is often quite small. Considering the small magnitude of $l_{i{+}1{,}i}$ and the additional shifting technique, the accuracy of our method will not get affected.  

\section{Experiments}
In this section, we first perform a numerical test to compare our method with SVD for matrices in different dimensions and batch sizes. Subsequently, we evaluate the effectiveness of the proposed methods in three computer vision applications: decorrelated BN, second-order vision transformer, and neural style transfer. The implementations details are referred to supplementary material. 

\subsection{Numerical Test}

\begin{figure*}
    \centering
    \includegraphics[width=0.99\linewidth]{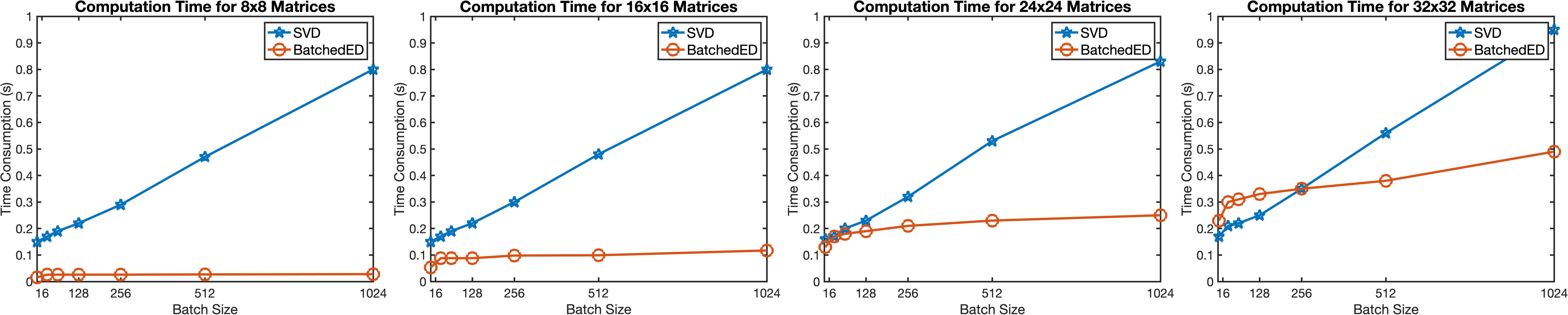}
    \caption{The speed comparison of our Batched ED against \textsc{torch.svd} for different batch sizes and matrix dimensions. Our implementation is more batch-friendly and the time cost does not vary much against different batch sizes. For matrices in small and moderate sizes, our method can be significantly faster than the Pytorch SVD.}
    \label{fig:numerical_test}
\end{figure*}

Fig.~\ref{fig:numerical_test} depicts the computational time of our Batched ED against the SVD for different matrix dimensions and batch sizes. The time cost of the SVD grows almost linearly with the batch size, while the time consumption of our Batched ED only has slight or mild changes against varying batch sizes. For matrices whose dimensions are smaller than $24$, our Batched ED is consistently faster than the SVD for any batch size. When the matrix dimension is $32$, our method is faster than the SVD from batch size $256$ on. The speed of our Batched ED is more advantageous for smaller matrix dimensions and larger batch sizes.





\subsection{Decorrelated BN}

\begin{table}[h]
    \centering
    \setlength{\tabcolsep}{1.5pt}
    \caption{Validation error of decorrelated BN on ResNet-18~\cite{he2016deep}. The results are reported based on $5$ runs, and we measure the time of the forward ED in a single step.}
    \label{tab:zca_norm}
     \resizebox{0.8\linewidth}{!}{
    \begin{tabular}{r|c|c|c|c|c|c|c}
    \toprule
        \multirow{2}*{Solver} & \multirow{2}*{Group} & \multirow{2}*{Size} & \multirow{2}*{Time (s)} &  \multicolumn{2}{c|}{CIFAR10} & \multicolumn{2}{c}{CIFAR100} 
        \\\cline{5-8}
        & & & & mean$\pm$std & min  & mean$\pm$std & min \\
        \hline
        SVD    &  \multirow{2}*{$16$}  & \multirow{2}*{$16{\times}4{\times}4$} &0.172 & $4.52{\pm}0.09$&4.33 &\textbf{21.24$\pm$0.17} & {20.99}\\ 
        Batched ED & & &\textbf{\textcolor{red}{0.006}} & \textbf{4.37$\pm$0.11} &\textbf{4.29} & 21.25$\pm$0.20 & \textbf{20.90}\\
        \hline
        SVD    &  \multirow{2}*{$8$}  & \multirow{2}*{$8{\times}8{\times}8$} &0.170 & $4.55{\pm}0.13$&4.34 & $21.32{\pm}0.31$&20.88\\ 
        Batched ED & & &\textbf{\textcolor{red}{0.016}} &\textbf{4.36$\pm$0.11} &\textbf{4.25} & \textbf{20.97$\pm$0.27} &\textbf{20.62}\\
        \hline
        SVD    &  \multirow{2}*{$4$}  & \multirow{2}*{$4{\times}16{\times}16$} &0.165 &$4.52{\pm}0.14$ &4.33 &$21.30{\pm}0.33$ & \textbf{20.86}\\ 
        Batched ED & & &\textbf{\textcolor{red}{0.075}} & \textbf{4.45$\pm$0.11} &\textbf{4.32} & \textbf{21.19$\pm$0.21} & 20.98\\
    \bottomrule
    \end{tabular}
    }
\end{table}

Following~\cite{song2022fast}, we first conduct an experiment on the task of ZCA whitening. In the whitening process, the inverse square root of the covaraince is multiplied with the feature as $(\mX\mX^{T})^{-\frac{1}{2}}\mX$ to eliminate the correlation between each dimension. We insert the ZCA whitening meta-layer into the ResNet-18~\cite{he2016deep} architecture and evaluate the validation error on CIFAR10 and CIFAR100~\cite{krizhevsky2009learning}. Table~\ref{tab:zca_norm} compares the performance of our Batched ED against the SVD. Depending on the number of groups, our method can be $2$X faster, $10$X faster, and even $28$X faster than the SVD. Furthermore, our method outperforms the SVD across all the metrics on CIFAR10. With CIFAR100, the performance is also on par.

\subsection{Second-order Vision Transformer}

\begin{table}[t]
    \centering
    \setlength{\tabcolsep}{1.5pt}
    \caption{Validation accuracy on ImageNet~\cite{deng2009imagenet} for the second-order vision transformer with different depths. Here $32$ and $36$ denote the spatial dimension of visual tokens. We report the time consumption of the forward ED in a single step. }
    \label{tab:2nd_transformer}
     \resizebox{0.7\linewidth}{!}{
    \begin{tabular}{r|c|c|c|c}
    \toprule
          \multirow{2}*{Solver} & \multirow{2}*{Size} & \multirow{2}*{Time (s)} &\multicolumn{2}{c}{Architecture}  \\ \cline{4-5}
          & & & So-ViT-7 & So-ViT-10 \\
         \hline
          SVD & \multirow{2}*{$768{\times}32{\times}32$} &0.767 &  76.01 / \textbf{93.10}  & \textbf{77.97} / \textbf{94.10}\\
         Batched ED & &\textbf{\textcolor{red}{0.431}} & \textbf{76.04} / 93.05 & 77.91 / 94.08  \\
         \hline
          SVD & \multirow{2}*{$768{\times}36{\times}36$} &0.835 & \textbf{76.10} / \textbf{93.14}  & {78.09} / 94.13\\
         Batched ED & &\textbf{\textcolor{red}{0.612}} & 76.07 / 93.10 & \textbf{78.11} / \textbf{94.19}\\
         \bottomrule
    \end{tabular}
    }
\end{table}

We turn to the experiment on the task of global covariance pooling for the Second-order Vision Transformer (So-ViT)~\cite{xie2021so}. To leverage the rich semantics embedded in the visual tokens, the covariance square root of the visual tokens $(\mX\mX^{T})^{\frac{1}{2}}$ are used to assist the classification task. Since the global covariance matrices are typically very ill-conditioned~\cite{song2021approximate}, this task poses a huge challenge to the stability of the ED algorithm. We choose the So-ViT architecture with different depths and validate the performance on ImageNet~\cite{deng2009imagenet}. As observed from Table~\ref{tab:2nd_transformer}, our Batched ED has the competitive performance against the standard SVD. Moreover, our method is about $44\%$ and $27\%$ faster than the SVD for covariance in different sizes. 


\begin{figure}[t]
    \centering
    \includegraphics[width=0.6\linewidth]{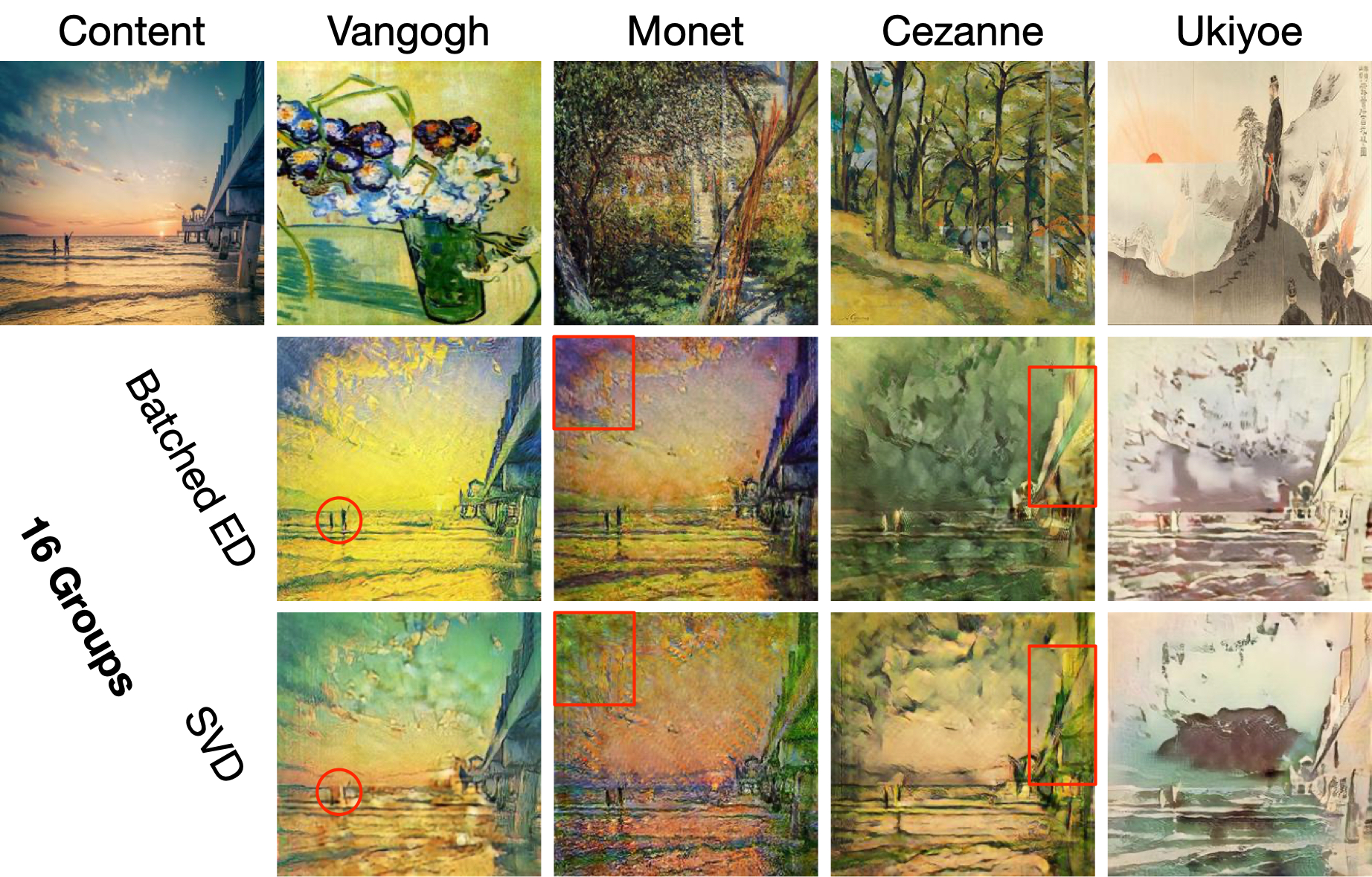}
    \caption{Exemplary visual comparison. The red circle/rectangular indicates the region with subtle details. In this example, our method generates sharper images with more coherent style information and less artifacts. Zoom in for a better view.}
    \label{fig:style_transfer_exp}
\end{figure}

\begin{table}[ht]
    \centering
    \setlength{\tabcolsep}{1.5pt}
    \caption{The LPIPS distance between the transferred image and the content image and the user preference (\%) on the Artworks~\cite{isola2017image} dataset. We report the time consumption of the forward ED that is conducted $10$ times to exchange the style and content feature at different network depths. The batch size is set to $4$.}
    \label{tab:style_transfer}
     \resizebox{0.75\linewidth}{!}{
    \begin{tabular}{r|c|c|c|c|c}
    \toprule
        Solver &Group & {Size} & {Time (s)} & {LPIPS~\cite{zhang2018perceptual} ($\uparrow$)} & Preference ($\uparrow$)
        \\
        \hline
        SVD    &  \multirow{2}*{$64$}  & \multirow{2}*{$256{\times}4{\times}4$} &3.146 &0.5776 & \textbf{48.25}\\ 
        Batched ED & & &\textbf{\textcolor{red}{0.089}}  &\textbf{0.5798} &47.75\\
        \hline
        SVD    &  \multirow{2}*{$32$}  & \multirow{2}*{$128{\times}8{\times}8$} &2.306 &\textbf{0.5722} & 47.75\\ 
        Batched ED & & & \textbf{\textcolor{red}{0.257}}  &0.5700 & \textbf{48.75}\\
        \hline
        SVD    &  \multirow{2}*{$16$}  & \multirow{2}*{$64{\times}16{\times}16$} &1.973 &0.5614 &46.25\\ 
        Batched ED & & &\textbf{\textcolor{red}{0.876}} & \textbf{0.5694} &\textbf{47.75}\\
    \bottomrule
    \end{tabular}
    }
\end{table}

\subsection{Universal Style Transfer}

Now we apply our Batched ED in the WCT for neural style transfer. Given the content feature $\mX_{c}$ and the style feature $\mX_{s}$, the WCT performs successive whitening ($(\mX_{c}\mX_{c})^{-\frac{1}{2}}\mX_{c}$) and coloring ($(\mX_{s}\mX_{s})^{\frac{1}{2}}\mX_{c}$) to transfer the target style. We follow~\cite{li2017universal,wang2020diversified} to use the LPIPS distance and the user preference as the evaluation metrics. Table~\ref{tab:style_transfer} presents the quantitative comparison with different groups. Our Batched ED achieves very competitive performance and predominates the speed. To give a concrete example, when the group number is $64$, our method is about $35$X faster than the default SVD. Fig.~\ref{fig:style_transfer_exp} displays the exemplary visual comparison. In this specific example, our Batched ED generates images with better visual appeal.

\begin{figure}[htbp]
    \centering
    \includegraphics[width=0.99\linewidth]{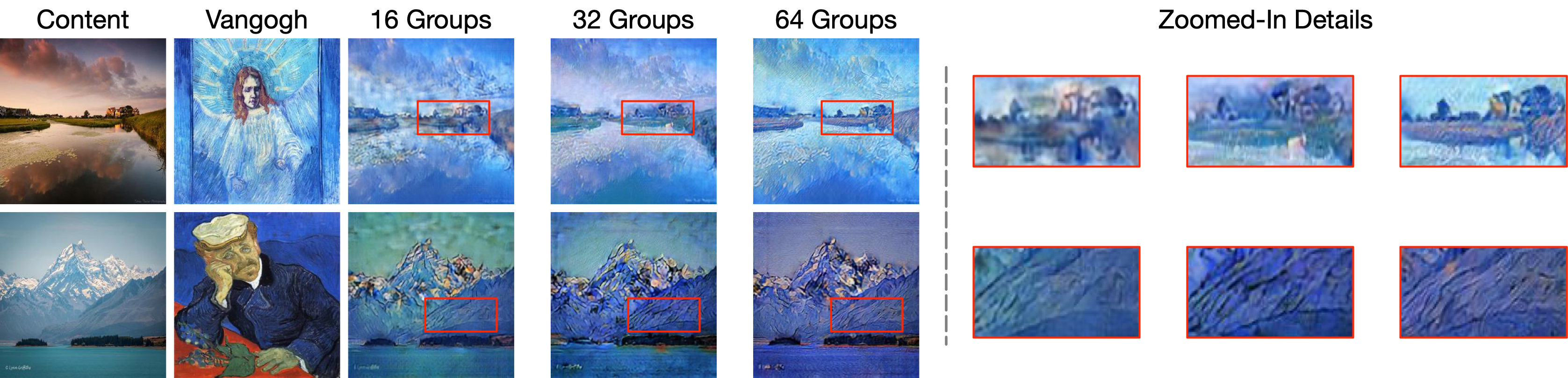}
    \caption{Visual illustration of the impact of groups. When more groups are used, the strength of the target style is increased and the details are better preserved.}
    \label{fig:style_transfer_group}
\end{figure}

Similar to the finding in~\cite{cho2019image}, we also observe that the number of groups has an impact on the extent of transferred style. As shown in Fig.~\ref{fig:style_transfer_group}, when more groups are used, the style in the transferred image becomes more distinguishable and the details are better preserved. Since the number of groups determines the number of divided channels and the covariance size, more groups correspond to smaller covariance and this might help to better capture the local structure. Despite this superficial conjecture, giving a more comprehensive and rigorous analysis is worth further research.

To sum up, our ED solver has demonstrated the superior batch efficiency for small matrices in various real-world experiments and numerical tests. The limitation on large matrices indicates the key difference: \emph{our method is more batch-efficient, while \textsc{torch.eig/svd} is more dimension-efficient.}
\section{Conclusion}

In this paper, we propose a batch-efficient QR-based ED algorithm dedicated for batched matrices which are common in the context of computer vision and deep learning. Aided by the proposed acceleration techniques, our solver is much faster than Pytorch SVD function for a mini-batch of small and medium matrices. Our method can directly benefit a wide range of computer vision applications and we showcase this merit in several applications of differentiable ED. Extensive experiments on visual recognition and image generation demonstrate that our method can also achieve very competitive performances.


\clearpage
%
%
\bibliographystyle{splncs04}
\bibliography{egbib}

\appendix

This document presents the detailed illustration of some used techniques (Sec.~\ref{sec:der}), the introduction of our experimental settings and implementation details (Sec.~\ref{sec:exp}), and the detailed analysis and discussion of our method (Sec.~\ref{sec:svd}). 

\section{Theoretical Deviation}
\label{sec:der}

\subsection{$2{\times}2$ Givens Rotation}
Our goal is to eliminate the sub-diagonal/super-diagonal entry of a $2{\times}2$ matrix. Consider a given vector $\mathbf{x}^{T}{=}{[}x_{1}{,}x_{2}{]}$. We can set the entries of Givens rotation as:
\begin{equation}
    \cos{\theta}=\frac{x_{1}}{\sqrt{x_{1}^2+x_{2}^2}}, \sin{\theta}=-\frac{x_{2}}{\sqrt{x_{1}^2+x_{2}^2}}
    \label{givens_vector_def}
\end{equation}
Then the rotation can eliminate the entry by:
\begin{equation}
    \mathbf{R}^{T}\mathbf{x}=\begin{bmatrix}
    \cos{\theta} & -\sin{\theta} \\
    \sin{\theta} & \cos{\theta} \\
    \end{bmatrix}\begin{bmatrix}
    x_{1} \\
    x_{2} \\
    \end{bmatrix}=\begin{bmatrix}
    \sqrt{x_{1}^2+x_{2}^2} \\
    0 \\
    \end{bmatrix}
    \label{givens_vector}
\end{equation}
For the simplicity concern, we use $c$ and $s$ to represent $\cos{\theta}$ and $\sin{\theta}$, respectively. Let us extend the vector $\mathbf{x}$ to a symmetric matrix $\mathbf{X}$. The Givens rotation is applied by an orthogonal similarity transform:
\begin{equation}
\begin{aligned}
    \mathbf{R}^{T}\mathbf{X}\mathbf{R} &= \begin{bmatrix}
    c & -s \\
    s & c \\
    \end{bmatrix}\begin{bmatrix}
    x_{1} & x_{2}\\
    x_{2} & x_{3}\\
    \end{bmatrix}\begin{bmatrix}
    c& s \\
    -s & c \\
    \end{bmatrix}\\
    &=\begin{bmatrix}
    x_{1}c-x_{2}s & x_{2}c-x_{3}s \\
    x_{1}s+x_{2}c & x_{2}s+x_{3}c \\
    \end{bmatrix}\begin{bmatrix}
    c& s \\
    -s & c \\
    \end{bmatrix}\\
    &=\begin{bmatrix}
    x_{1}c^2{-}2x_{2}cs{+}x_{3}s^2 & x_{2}(c^2{-}s^2){+}{(}x_{1}{-}x_{3}{)}cs \\
    x_{2}{(}c^2{-}s^2{)}{+}{(}x_{1}{-}x_{3}{)}cs  & x_{1}s^2{+}2x_{2}cs{+}x_{3}c^2 \\
    \end{bmatrix}\\
\end{aligned}
\label{givens_matrix}
\end{equation}
As can be seen, the symmetric and orthogonal form still manifests after rotation. From~\cref{givens_vector_def} and~\cref{givens_vector}, we have $cx_{2}{+}sx_{1}{=}0$. Injecting this relation into~\cref{givens_matrix} leads to the re-formulation:
\begin{equation}
   \begin{bmatrix}
   x_{1}c^2{-}2x_{2}cs{+}x_{3}s^2 & -x_{2}s^2-x_{3}cs\\
   -x_{2}s^2-x_{3}cs & x_{1}s^2{+}2x_{2}cs{+}x_{3}c^2\\
   \end{bmatrix}
\end{equation}
The magnitude of sub-diagonal entries gets smaller. A series of such Givens rotations moving along the diagonal form the orthogonal matrix $\mathbf{Q}_{k}$ for a QR iteration:
\begin{equation}
    \mathbf{Q}_{k}=\mathbf{R}_{0:2}\mathbf{R}_{1:3}\dots\mathbf{R}_{N-2:N}
\end{equation}
where the sub-script of $\mathbf{R}$ denotes the region where the the rotation is applied. Notice that the successive Givens rotations still keep the tri-diagonal form of the matrix but gradually reduce the strength of super-diagonal entries. This accounts for why the QR iterations can transform a tri-diagonal matrix into a diagonal one. 

\subsection{Convergence of QR iteration}

We give the proof for the theorem about convergence speed of QR iterations. 

\setcounter{thm}{1}
\begin{thm}[Convergence of QR iteration]
Let $\mathbf{T}$ be the positive definite tri-diagonal matrix with the eigendecomposition $\mathbf{Q}\mathbf{\Lambda}\mathbf{Q}^{T}$ and assume $\mathbf{Q}^{T}$ can be LU decomposed. Then the QR iteration of $\mathbf{T}$ will converge to $\mathbf{\Lambda}$.
\end{thm}
\begin{proof}
Since we have $\mathbf{T}{=}\mathbf{Q}\mathbf{\Lambda}\mathbf{Q}^{T}$, then:
\begin{equation}
    \mathbf{T}^{k}=\mathbf{Q}\mathbf{\Lambda}^{k}\mathbf{Q}^{T}=(\mathbf{Q}_{0}\dots\mathbf{Q}_{k})(\mathbf{R}_{k}\dots\mathbf{R}_{0})
\end{equation}
By assuming $\mathbf{Q}^{T}{=}\mathbf{L}\mathbf{U}$, this equation can be written as:
\begin{equation}
\begin{gathered}
  \mathbf{Q}\mathbf{\Lambda}^{k}\mathbf{L}\mathbf{U}= (\mathbf{Q}_{0}\dots\mathbf{Q}_{k})(\mathbf{R}_{k}\dots\mathbf{R}_{0}) \\
  \mathbf{Q}\mathbf{\Lambda}^{k}\mathbf{L}\mathbf{\Lambda}^{-k}=(\mathbf{Q}_{0}\dots\mathbf{Q}_{k})(\mathbf{R}_{k}\dots\mathbf{R}_{0})\mathbf{U}^{-1}\mathbf{\Lambda}^{-k}
\end{gathered}
\end{equation}
For $\mathbf{\Lambda}^{k}\mathbf{L}\mathbf{\Lambda}^{-k}$, its entry is defined by:
\begin{equation}
    (\mathbf{\Lambda}^{k}\mathbf{L}\mathbf{\Lambda}^{-k})_{i,j}=\begin{cases}
    l_{i,j}(\frac{\lambda_{i}}{\lambda_{j}})^{k} & i>j \\
    1 & i=j\\
    0 & otherwise
    \end{cases}
\end{equation}
When $k{\rightarrow}\infty$, we have $(\frac{\lambda_{i}}{\lambda_{j}})^{k}{\rightarrow}0$ and $\mathbf{\Lambda}^{k}\mathbf{L}\mathbf{\Lambda}^{-k}{\rightarrow}\mathbf{I}$. Due to the uniqueness of the QR factorization, we also have $\mathbf{Q}_{0}{\dots}\mathbf{Q}_{k}{\rightarrow}\mathbf{Q}$ and $\mathbf{R}_{k}{\dots}\mathbf{R}_{0}\mathbf{U}^{-1}\mathbf{\Lambda}^{-k}{\rightarrow}\mathbf{I}$. Then the QR iterations can be formulated as:
\begin{equation}
   (\mathbf{Q}_{k}^{T}\dots\mathbf{Q}_{0}^{T})\mathbf{T}(\mathbf{Q}_{0}\dots\mathbf{Q}_{k}){\rightarrow}\mathbf{Q}^{T}\mathbf{T}\mathbf{Q}=\mathbf{\Lambda}
\end{equation}
As seen above, the QR iteration converges to the eigenvalue. 
\end{proof}

This theorem implies that the convergence speed of QR iterations is actually dependent on the adjacent eigenvalue ratio $\nicefrac{\lambda_{i}}{\lambda_{j}}$ for $i{>}j$. 

\subsection{Wilkinson Shift}

The Wilkinson shift denotes extracting the two eigenvalues from the right bottom $2{\times}2$ block of a matrix and uses them as the shift coefficients. This can be also accomplished by Givens rotation. Consider the general form of the orthogonal transform by Givens rotation in~\cref{givens_matrix}. Setting the off-diagonal entries to zero leads to the linear equations:
\begin{equation}
    \begin{cases}
    (x_{1}-x_{3})cs+x_{2}(c^2-s^2) = 0 \\
    c^2 + s^2 = 1\\
    \end{cases}
    \label{givens_equ}
\end{equation}
Let we define two variables by:
\begin{equation}
    m=\frac{x_{1}-x_{3}}{2x_{2}},\ n=\frac{s}{c}=\tan{\theta}
\end{equation}
Then \cref{givens_equ} is equivalent to:
\begin{equation}
    n^2 - 2mn - 1=0
\end{equation}
The above equation has two roots that are defined by
\begin{equation}
    n= m \pm \sqrt{1+m^2}
\end{equation}
We select the smaller one to ensure that the rotation angle $\theta$ is within 45 degrees. Then the entries of the rotation are given by:
\begin{equation}
    c=\frac{1}{\sqrt{1+m^2}},\ s=cn
\end{equation}
The two eigenvalues are derived and used as the shifts.

\subsection{Implicit Q Theorem}
In the paper, we present the following implicit Q theorem without proof.
\setcounter{thm}{2}
\begin{thm}[Implicit Q Theorem]
Let $\mathbf{B}$ be an upper Hessenberg and only have positive elements on its first sub-diagonal. Assume there exists a unitary transform $\mathbf{Q}^{H}\mathbf{A}\mathbf{Q}{=}\mathbf{B}$. Then $\mathbf{Q}$ and $\mathbf{B}$ are uniquely determined by $\mathbf{A}$ and the first column of $\mathbf{Q}$.
\end{thm}

Now we give a short proof and illustrate why the theorem cannot be directly applied in our case.

\begin{proof}
 Since the QR iteration is a unitary transform, we can write $\mathbf{Q}^{H}\mathbf{A}\mathbf{Q}{=}\mathbf{B}$ as:
 \begin{equation}
     \mathbf{A}\mathbf{Q}=\mathbf{Q}\mathbf{B}
     \label{implicit_q_aq_qb}
 \end{equation}
If we represent $\mathbf{Q}$ by a vector of columns $\mathbf{Q}{=}{[}\mathbf{q}_{0}{,}{\dots}{,}\mathbf{q}_{n-1}{]}$,~\cref{implicit_q_aq_qb} can be re-written as:
\begin{equation}
    \mathbf{A}{[}\mathbf{q}_{0}{,}{\dots}{,}\mathbf{q}_{n-1}{]}={[}\mathbf{q}_{0}{,}{\dots}{,}\mathbf{q}_{n-1}{]}\mathbf{B}
\end{equation}
Recall that $\mathbf{B}$ is a tri-diagonal matrix and only has non-zero entries at $b_{i-1,i}$, $b_{i,i}$, and $b_{i+1,i}$. Relying on this property, we have:
\begin{equation}
\begin{gathered}
    \mathbf{A}\mathbf{q}_{i+1} = b_{i-1,i}\mathbf{q}_{i-1} + b_{i,i}\mathbf{q}_{i}+ b_{i+1,i}\mathbf{q}_{i+1} \\
    \mathbf{q}_{i+1} = \frac{\mathbf{A}\mathbf{q}_{i+1} - b_{i-1,i}\mathbf{q}_{i-1} - b_{i,i}\mathbf{q}_{i}}{b_{i+1,i}}
    \label{implicit_q_b}
\end{gathered}
\end{equation}
Since $\mathbf{q}$ is orthogonal, \emph{i.e.,} $\mathbf{q}^{T}\mathbf{q}{=}\mathbf{I}$, we have:
\begin{equation}
    \begin{gathered}
        b_{i+1,i}=||\mathbf{A}\mathbf{q}_{i+1} - b_{i-1,i}\mathbf{q}_{i-1} - b_{i,i}\mathbf{q}_{i}||_{2}\\
        \mathbf{q}_{i+1} = \frac{\mathbf{A}\mathbf{q}_{i+1} - b_{i-1,i}\mathbf{q}_{i-1} - b_{i,i}\mathbf{q}_{i}}{||\mathbf{A}\mathbf{q}_{i+1} - b_{i-1,i}\mathbf{q}_{i-1} - b_{i,i}\mathbf{q}_{i}||_{2}}
    \end{gathered}
\end{equation}
As indicated above, each column of $\mathbf{Q}$ and the sub-diagonal entries of $\mathbf{B}$ can be uniquely computed by the previous columns. 
\end{proof}

This theorem presents an algorithm that could greatly simplify the QR iterations without explicit Givens rotations. However, as can be seen from~\cref{implicit_q_b}, the theorem relies on the assumption that the sub-diagonal entry $b_{i+1,i}$ is non-zero. In our case, any Givens rotation aims at zeroing out the sub-diagonal entry $b_{i+1,i}$. As a consequence, $b_{i+1,i}$ are very small and even can be zero after rotation, which violates the assumption. This is more serious for batched matrices, as more matrices could amplify the probability. Directly applying the theorem could introduce large round-off errors and may cause data overflow.

Nonetheless, this theorem implies that the $i$-th column of $\mathbf{Q}$ only depends on the previous two columns of $\mathbf{Q}$ and $\mathbf{B}$, but not on the columns after $i$-th column. This shows that the $i$-th Givens rotation will only affect part of $\mathbf{Q}$. Therefore, we propose our economic eigenvector calculation to involve part of the matrix for each rotation.

\section{Experimental Settings}
\label{sec:exp}

\subsection{Implementation Details} 
All the source codes are implemented in Pytorch 1.7.0 with the self-contained CUDA wrapper. Older versions till 1.0.0 should be also compatible. We compare our method with the function \textsc{torch.svd}, which calls the LAPACK's SVD \textit{gesdd} routine that uses the divide-and-conquer strategy to solve the eigenvalue problem. Some other functions, such as \textsc{torch.symeig} and \textsc{torch.eig}, can be also adopted to perform the ED. However, we empirically found that they are not as numerically stable as \textsc{torch.svd} and often cause the network to fail in converging. We thus only compare our method with the function \textsc{torch.svd}. All the numerical tests are conducted on a workstation equipped with a GeForce GTX 1080Ti GPU and a 6-core Intel(R) Xeon(R) GPU @ 2.20GHz. The computational time is measured based on $10,000$ randomly generated matrices. We note that if a low-level programming language (\emph{e.g.,} CUDA C++) is used to implement our algorithm, the speed might get further improved.

For our method, the threshold for the dimension reduction is set as $1e{-}5$. Our process of batched Givens diagonalization lasts $n$ iterations, where each iteration consists of two sequential QR iterations with Wilkinson shifts. We use the techniques implemented in~\cite{song2021approximate} for the backward gradient computation.

\subsection{Decorrelated BN} 

To perform the ZCA whitening, given the reshaped feature map $\mathbf{X}{\in}\mathrm{R}^{C{\times}BHW}$, the covaraince of the feature is first computed as:
\begin{equation}
    \mathbf{A}=(\mathbf{X}-\mu)(\mathbf{X}-\mu)^{T}+\epsilon\mathbf{I}
\end{equation}
where $\mathbf{A}{\in}\mathrm{R}^{C{\times}C}$, $\mu$ is the mean of $\mathbf{X}$, and $\epsilon$ is a small positive constant to ensure the positive-definitiveness of $\mathbf{A}$. Afterwards, the inverse square root of the covaraince is applied to whiten the feature:
\begin{equation}
    \mathbf{X}_{whitened}=\mathbf{A}^{-\frac{1}{2}}\mathbf{X}
\end{equation}
Compared with the BN, the whitened feature map $\mathbf{X}_{whitened}$ further eliminates the data correlation between the features. The statistics $\mu$ and $\mathbf{A}^{-\frac{1}{2}}$ during the training phase are stored and used in the inference stage.

In the practical implementation, one often split the feature $\mathbf{X}$ into multiple groups in the channel dimension and attain a mini-batch of matrices. This division allows each group to have own training statistics, which could improve the stability and generalization performance. This is particularly helpful in the regime of small batch sizes~\cite{huang2018decorrelated}. Let $G$ denote the group numbers. Then the split can be formally defined by:
\begin{equation}
    \mathbf{X}\in\mathrm{R}^{C\times BHW} \rightarrow \mathbf{X}\in\mathrm{R}^{\frac{C}{G}\times G\times BHW}
\end{equation}
The covariance is changed accordingly as:
\begin{equation}
    \mathbf{A}\in\mathrm{R}^{C\times C} \rightarrow \mathbf{A}\in\mathrm{R}^{\frac{C}{G}\times G\times G}
\end{equation}
As indicated above, the covariance becomes a mini-batch of matrices. The calculation raises the need of our batch-friendly ED algorithm. 

\begin{figure}[htbp]
    \centering
    \includegraphics[width=0.3\linewidth]{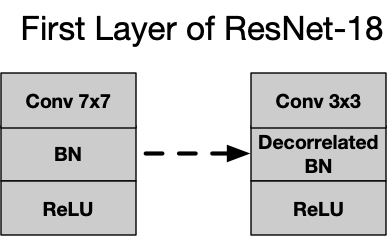}
    \caption{The detailed architecture changes of ResNet-18 after replacing the BN layer with the ZCA whitening meta-layer. Following~\cite{wang2021robust}, we reduce both the kernel size and the strides of the first convolution layer. The rest blocks of the model are not modified.}
    \label{fig:zca_arch}
\end{figure}

As depicted in Fig.~\ref{fig:zca_arch}, we insert the decorrelated BN layer after the first convolution layer of the ResNet-18 architecture. For the calculation of backward gradients, we use the Taylor polynomial for gradient approximation~\cite{song2021approximate,wang2021robust}. The degree of the Taylor polynomial is set to $9$. For the other settings, we follow the experiments in~\cite{wang2021robust}.

\subsection{Second-order Vision Transformer} 

\begin{figure}[htbp]
    \centering
    \includegraphics[width=0.6\linewidth]{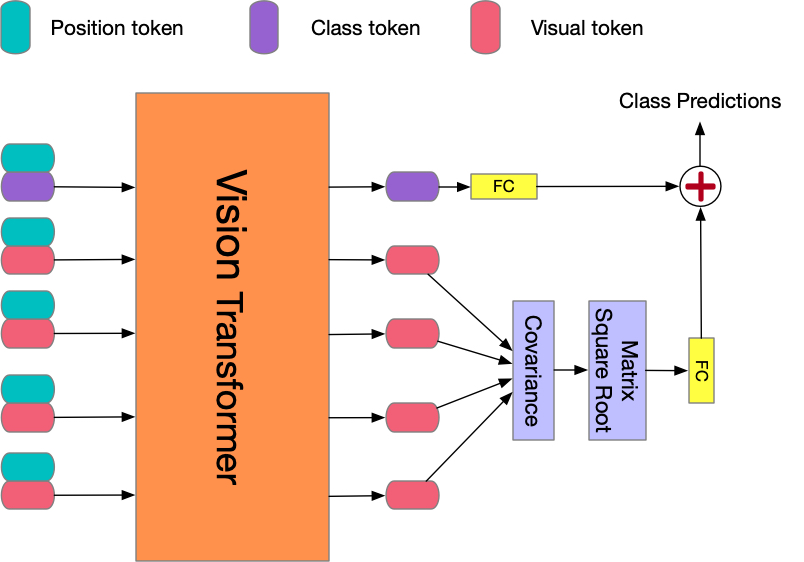}
    \caption{Scheme of the second-order vision transformer~\cite{xie2021so}. The covariance square root of the visual token is utilized to assist the classification task, which removes the need of pre-training on ultra-large-scale datasets. }
    \label{fig:2nd_transformer_arch}
\end{figure}

Fig.~\ref{fig:2nd_transformer_arch} depicts the architectural overview of the second-order vision transformer. In the ordinary vision transformer~\cite{dosovitskiy2020image}, only the class token is used to output the class predictions, which results in the need of pre-training on ultra-large-scale datasets. In the second-order vision transformer, the covariance square root of the visual tokens are utilized to assist the classification task. The process can be formulated as:
\begin{equation}
    y={\rm FC}(c) + {\rm FC} ((\mathbf{X}\mathbf{X}^{T})^{\frac{1}{2}})
\end{equation}
where $c$ is the class token, $\mathbf{X}$ denotes the visual token, and $y$ is the final class predictions. Equipped with the covariance pooling layer, the So-ViT model is free of pre-training and can achieve comparable performances with CNNs even if trained from scratch.

For training transformer architectures, people often use some advanced mixed-precision training techniques such as NVIDIA Apex. Due to the use of low-precision weights and data, these techniques can not only accelerate the training process, but also can reduce the risk of gradient explosion. However, as pointed out in~\cite{song2021approximate}, the low precision often poses a challenge to the SVD. As the SVD often needs double precision to attain the effective numerical representation of the eigenvalues, using a low precision (\emph{i.e.,} float or half) can cause the network to fail in converging. To avoid this issue, we first train the model using Newton-Schulz iteration that computes the approximate matrix square root for the first $50$ epochs. Then we switch to the SVD or our Batched ED and continue the training. This hybrid approach can avoid the non-convergence at the beginning of training.

We set the spatial dimension of the visual tokens to $32{\times}32$ and $36{\times}36$ in our paper. The batch size is set as $768$. For the other experimental settings, we follow~\cite{xie2021so}. 

\subsection{Universal Style Transfer}

\begin{table*}[htbp]
    \centering
    \setlength{\tabcolsep}{1.5pt}
    \caption{The LPIPS distance and the user preference ($\%$) on each sub-set of the Artworks~\cite{isola2017image} dataset. We report the time consumption of the forward ED process that is conducted $10$ times to exchange the style and content feature at different network depths.}
    \label{tab:style_transfer_full}
     \resizebox{0.99\linewidth}{!}{
    \begin{tabular}{r|c|c|c|c|c|c|c|c|c|c|c}
    \toprule
        \multirow{2}*{Solver} & \multirow{2}*{Group} & \multirow{2}*{Size} & \multirow{2}*{Time (s)} &  \multicolumn{4}{c|}{LPIPS~\cite{zhang2018perceptual} ($\uparrow$)} & \multicolumn{4}{c}{Preference ($\uparrow$)} 
        \\\cline{5-12}
        & & & & Vangogh & Monet & Cezanne & Ukiyoe & Vangogh & Monet & Cezanne & Ukiyoe \\
        \hline
        SVD    &  \multirow{2}*{$64$}  & \multirow{2}*{$256{\times}4{\times}4$} &3.146 &\textbf{0.5448} &\textbf{0.5317} &0.6035 &0.6306 &44 &47 &\textbf{49} &\textbf{53} \\ 
        Batched ED & & &\textbf{\textcolor{red}{0.089}} &0.5346 &0.5027 &\textbf{0.6229} &\textbf{0.6589}& \textbf{52}&\textbf{49} &45 & 45\\
        \hline
        SVD    &  \multirow{2}*{$32$}  & \multirow{2}*{$128{\times}8{\times}8$} &2.306 &\textbf{0.5298} &0.5127 & 0.5751 &\textbf{0.6713} & 44&\textbf{47} &\textbf{50} &\textbf{50}\\ 
        Batched ED & & & \textbf{\textcolor{red}{0.257}} &0.5096 &\textbf{0.5258} &\textbf{0.6208} &0.6239 & \textbf{55}& 45 &48 &47\\
        \hline
        SVD    &  \multirow{2}*{$16$}  & \multirow{2}*{$64{\times}16{\times}16$} &1.973 & 0.4987& \textbf{0.5257} &0.5882 & \textbf{0.6630} &41 &\textbf{48} & 45 & \textbf{51}\\ 
        Batched ED & & &\textbf{\textcolor{red}{0.876}} & \textbf{0.5085} &0.5151 & \textbf{0.6041} &0.6498 &\textbf{57} &43 & \textbf{49} & 42\\
    \bottomrule
    \end{tabular}
    }
\end{table*}

\begin{figure}
    \centering
    \includegraphics[width=0.8\linewidth]{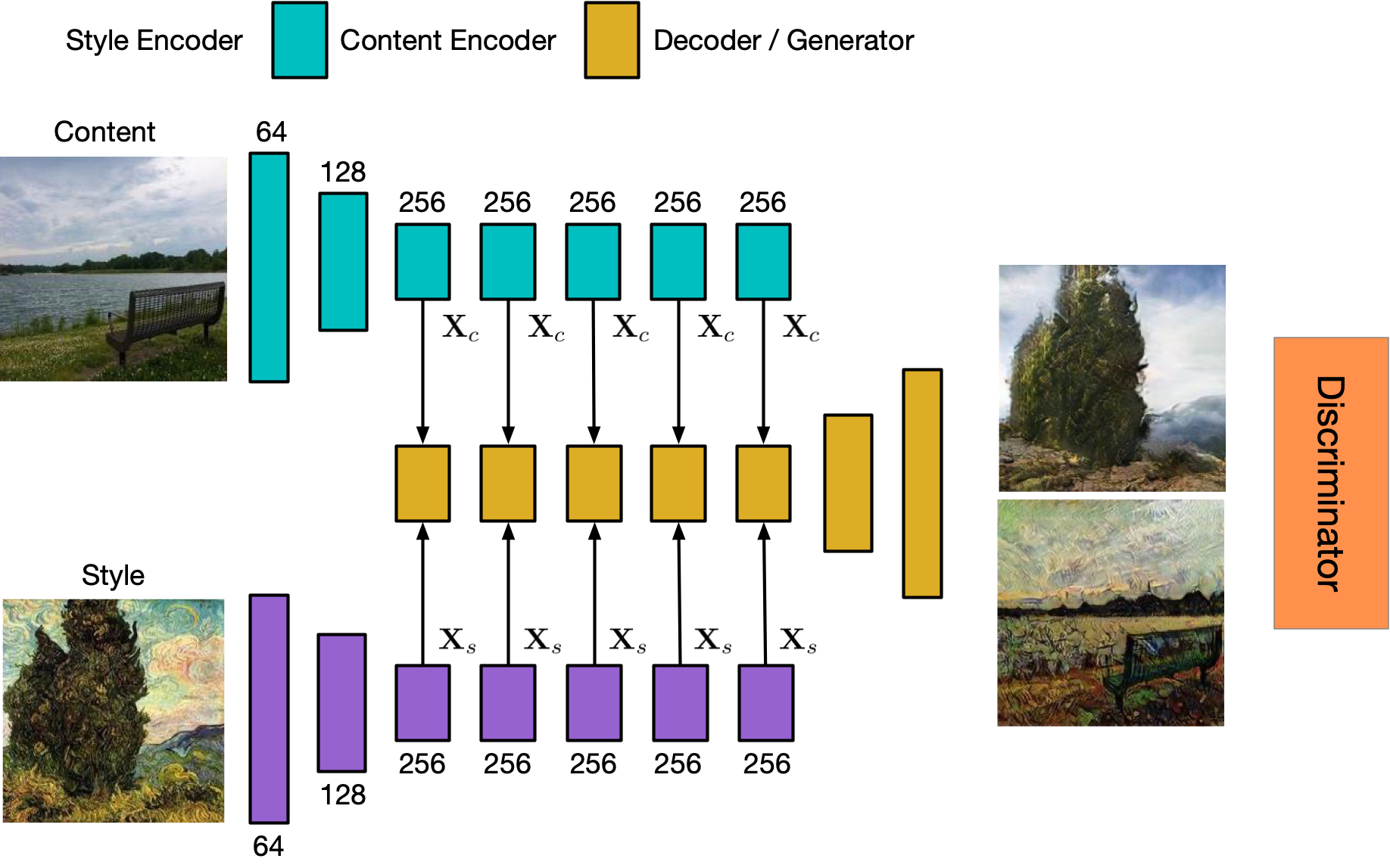}
    \caption{The architecture overview of our model for neural style transfer. Two encoders take input of the style and content image respectively, and generate the multi-scale content/style features. A decoder is applied to absorb the feature and perform the WCT process at $5$ different scales, which outputs a pair of images that exchange the styles. Finally, a discriminator is further adopted to tell apart the authenticity of the images. }
    \label{fig:style_transfer_arch}
\end{figure}

Following~\cite{wang2021robust}, we adopt the WCT process in the network architecture proposed in~\cite{cho2019image} for the universal style transfer. Fig.~\ref{fig:style_transfer_arch} displays the overview of the model. The WCT performs successive whitening and coloring transform on the content and style feature. Consider the content feature $\mathbf{X}_{c}{\in}\mathrm{R}^{B{\times}C{\times}H{\times}W}$ and the style feature $\mathbf{X}_{s}{\in}\mathrm{R}^{B{\times}C{\times}H{\times}W}$. We first divide the features into groups and reshape them as:
\begin{equation}
\begin{gathered}
     \mathbf{X}_{c}{\in}\mathrm{R}^{B{\times}C{\times}H{\times}W}{\rightarrow} \mathbf{X}_{c}{\in}\mathrm{R}^{\frac{BC}{G}{\times}G{\times}HW} \\
     \mathbf{X}_{s}{\in}\mathrm{R}^{B{\times}C{\times}H{\times}W}{\rightarrow} \mathbf{X}_{s}{\in}\mathrm{R}^{\frac{BC}{G}{\times}G{\times}HW} \\
\end{gathered}
\end{equation}
where $G$ denotes the group number. Subsequently, we remove the style information from the content feature as:
\begin{equation}
    \begin{gathered}
    \mathbf{X}_{c}^{whitened} = \Big((\mathbf{X}_{c}-\mu(\mathbf{X}_{c}))(\mathbf{X}_{c}-\mu(\mathbf{X}_{c}))^{T}\Big)^{-\frac{1}{2}}\mathbf{X}_{c}
    \end{gathered}
\end{equation}
Then we extract the desired style information from the style feature $\mathbf{X}_{s}$ and transfer it to the whitened content feature:
\begin{equation}
    \mathbf{X}_{c}^{colored} = \Big((\mathbf{X}_{s}-\mu(\mathbf{X}_{s}))(\mathbf{X}_{s}-\mu(\mathbf{X}_{s}))^{T}\Big)^{\frac{1}{2}}\mathbf{X}_{c}^{whitened}
\end{equation}
The resultant feature $\mathbf{X}_{c}^{colored}$ is compensated with the mean of style feature and combined with the original content feature:
\begin{equation}
    \mathbf{X} = \alpha (\mathbf{X}_{c}^{colored}+\mu(\mathbf{X}_{s})) + (1-\alpha)\mathbf{X}_{c} 
\end{equation}
where $\alpha$ is a weight bounded in $[0,1]$ to control the strength of style transfer. Finally, we feed the resultant feature $\mathbf{X}$ into the decoder to generate the realistic image where the style is transferred. 

For the loss functions, we follow~\cite{cho2019image} and use the cycle-consistent reconstruction loss in both the latent and the pixel space. The image is resized to the resolution of $216{\times}216$ before passing to the network, and the model is trained for $100,000$ iterations. The batch size is set to $4$.

Table~\ref{tab:style_transfer_full} displays the detailed quantitative evaluation. As suggested in~\cite{li2017universal,wang2020diversified}, we use the metrics LPIPS score between each pair of transferred image and the content image as well as the user preference. For the user study, we randomly select $100$ images from each dataset and ask $20$ volunteers to vote for the image that characterizes more the style information. In certain cases where the volunteer thinks neither of the two generated images correctly carries the style, he/she can abstain and does not vote for any one. 

\section{Detailed Analysis and Comparison}
\label{sec:svd}

\subsection{Speed Comparison against EIG}

\begin{table}[htbp]
    \centering
    \caption{Speed comparison against \textsc{torch.eig}. The results (ms) are reported in the format of our BatchedED / \textsc{torch.eig}.}
    \resizebox{0.8\linewidth}{!}{
    \begin{tabular}{c|c|c|c|c}
    \hline
         \diagbox{Batch Size}{Matrix Dim} & 4 & 8 & 16 & 24  \\
    \hline
         1 & \textbf{4}/5 & \textbf{8}/\textbf{8} & 53/\textbf{25} & 98/\textbf{59} \\
         4 & \textbf{6}/7 & 13/\textbf{9}& 75/\textbf{44} & 113/\textbf{68}\\
         16 & \textbf{7}/10 & 18/\textbf{10} & 88/\textbf{69} & 146/\textbf{92} \\
         64 & \textbf{8}/40 & \textbf{24}/49 & \textbf{90}/130 & \textbf{170}/210\\
         256 & \textbf{9}/160 & \textbf{26}/163 & \textbf{98}/219 & \textbf{191}/343\\
         1024 & \textbf{9}/610 & \textbf{28}/625 & \textbf{117}/749 & \textbf{270}/890\\
    \hline
    \end{tabular}
    }
    \label{tab:eig_comparison}
\end{table}

Though \textsc{torch.eig} is not numerically stable, its speed is usually faster than \textsc{torch.svd}. To ensure a fair comparison, we also compare our ED solver against \textsc{torch.eig} in Table~\ref{tab:eig_comparison}. Our method is more efficient in the regime of large batch sizes. When the batch size is small, our method also has comparable speed. The time cost of our BatchedED grows cubically versus matrix dimensions, whereas the cost of \textsc{torch.eig} drastically increases when the batch size increases. This demonstrates that our method is more batch-efficient, while \textsc{torch.eig/svd} is more dimension-efficient.

\subsection{Error Evaluation}

\begin{figure}[htbp]
    \centering
    \includegraphics[width=0.6\linewidth]{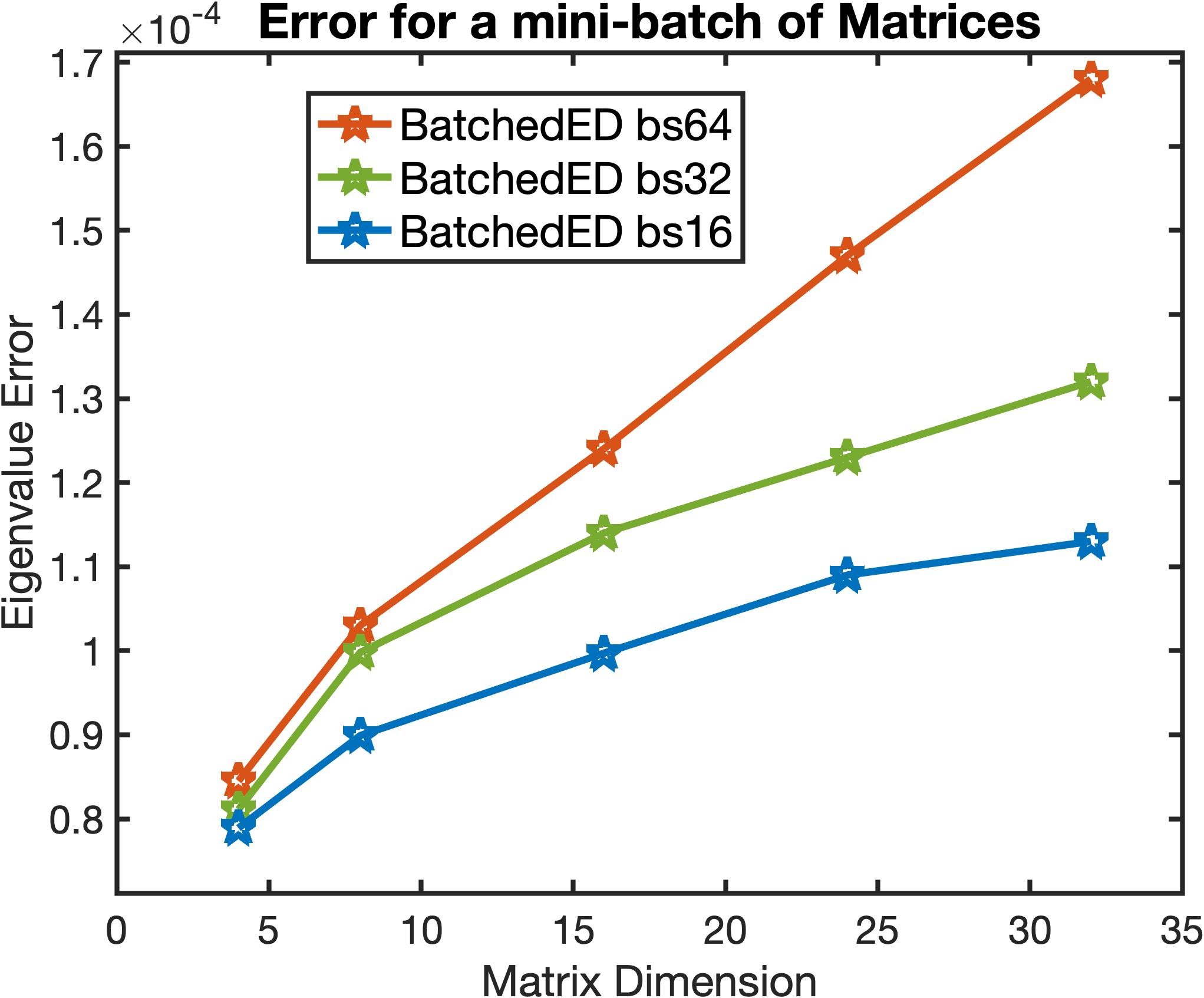}
    \caption{The error of eigenvalue estimation $||\mathbf{\Lambda}-\hat{\mathbf{\Lambda}}||_{\rm F}$ for a mini-batch of matrices where $\hat{\mathbf{\Lambda}}$ denotes the ground truth eigenvalue computed by SVD, and $\mathbf{\Lambda}$ represents the eigenvalue of our Batched ED.}
    \label{fig:error_dim}
\end{figure}

Fig.~\ref{fig:error_dim} displays the computation error of the eigenvalue for a mini-batch of matrices in different dimensions. When the batch size or matrix dimension increases, the error increases accordingly. However, the overall error is at an acceptance level for (${<}2e{-}4$). The computer vision experiments in the paper also demonstrate that such a small error will not affect the performance.

\subsection{Memory Usage Comparison}

\begin{figure}[htbp]
    \centering
    \includegraphics[width=0.6\linewidth]{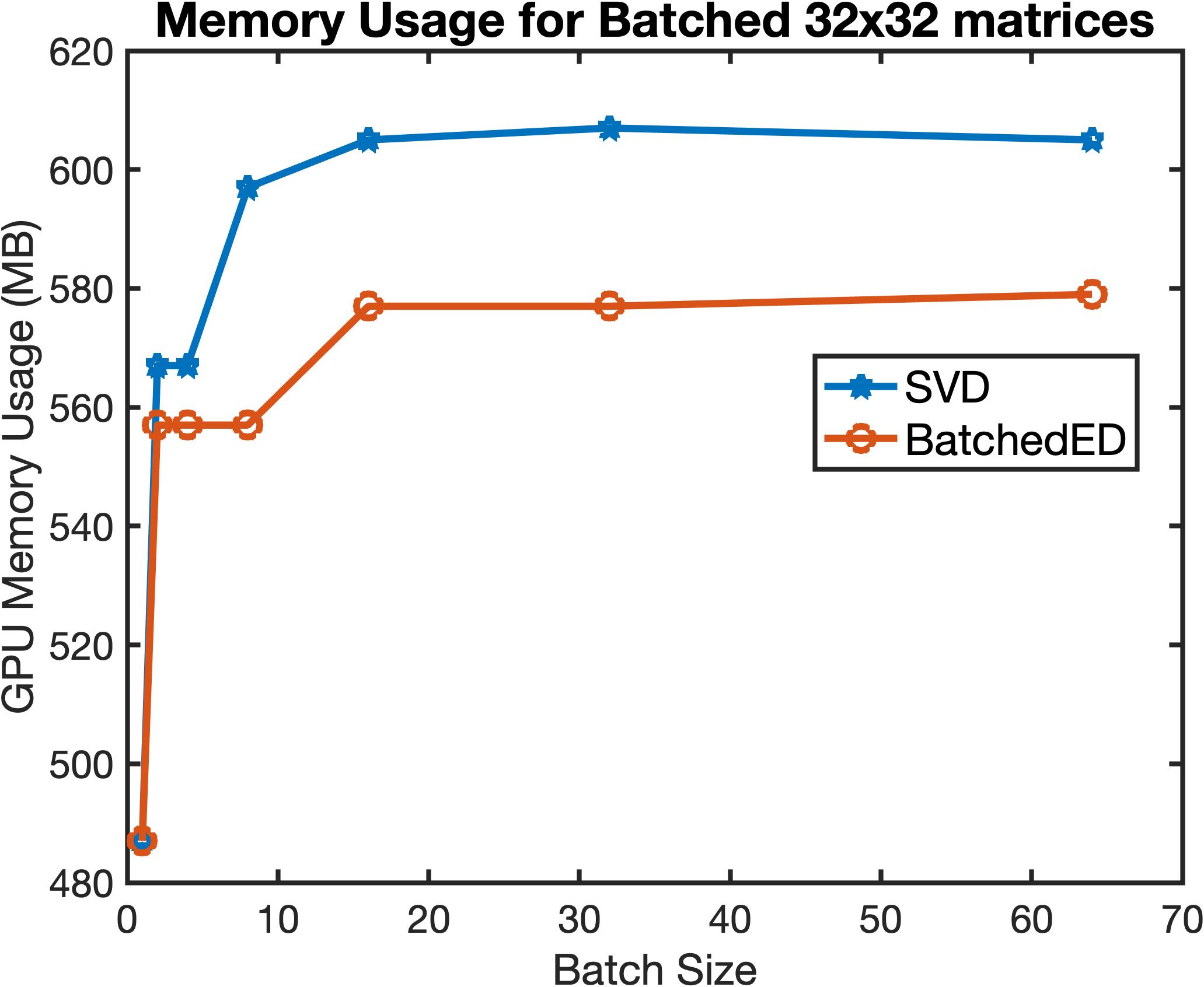}
    \caption{The comparison of GPU memory for a mini-batch of $32{\times}32$ matrices. }
    \label{fig:gpu_usage}
\end{figure}

Fig.~\ref{fig:gpu_usage} displays the GPU memory usage (MB) of performing the ED for a mini-batch of $32{\times}32$ matrices. For a single matrix, both methods consume almost the same memory. When the input scales to batched matrices, our Batched ED uses slightly less memory than the default LAPACK's SVD routine.

\subsection{Average Reduction Times}
As discussed in the paper, the speed of our Batched ED is greatly improved by the reduction times $r$ for the matrix shrinkage. More specifically, the time complexity is reduced by $O{(}{-}256n{(}1{+}r){)}$ for deriving eigenvalues and by $O{(}{-}{(}2r{+}1{)}n^{4}{)}$ for eigenvectors. Since our double Wilkinson shifts guarantee that the last two diagonal entries converge to zero quickly, \emph{i.e.,} $\frac{\lambda_{i}-\mu}{\lambda_{j}-\mu}{=}\infty$, the matrix dimension is very likely to shrink by one every QR iteration or every other QR iteration. According to our observations, the average reduction times $r$ is mainly in the range ${[}\frac{n}{3},\frac{3n}{4}{]}$.

\subsection{Why Batched ED Outperforms SVD}
In some experiments of the paper, our Batched ED can even achieve slightly better performances than the SVD. We think it is related to the implementation technique and the data precision. Due to the computation of secular equations, the divide-and-conquer strategy used in the \textsc{torch.svd} naturally requires a higher precision than the QR iterations of our method. Solving secular equations needs to perform the rational osculatory interpolation and it is more likely to trigger round-off errors when using a low data precision. In the regime of single-precision or half-precision, our QR-based Batched ED algorithm might have a slight advantage. 

\subsection{Limitation of Our Method}
\begin{figure}[htbp]
    \centering
    \includegraphics[width=0.6\linewidth]{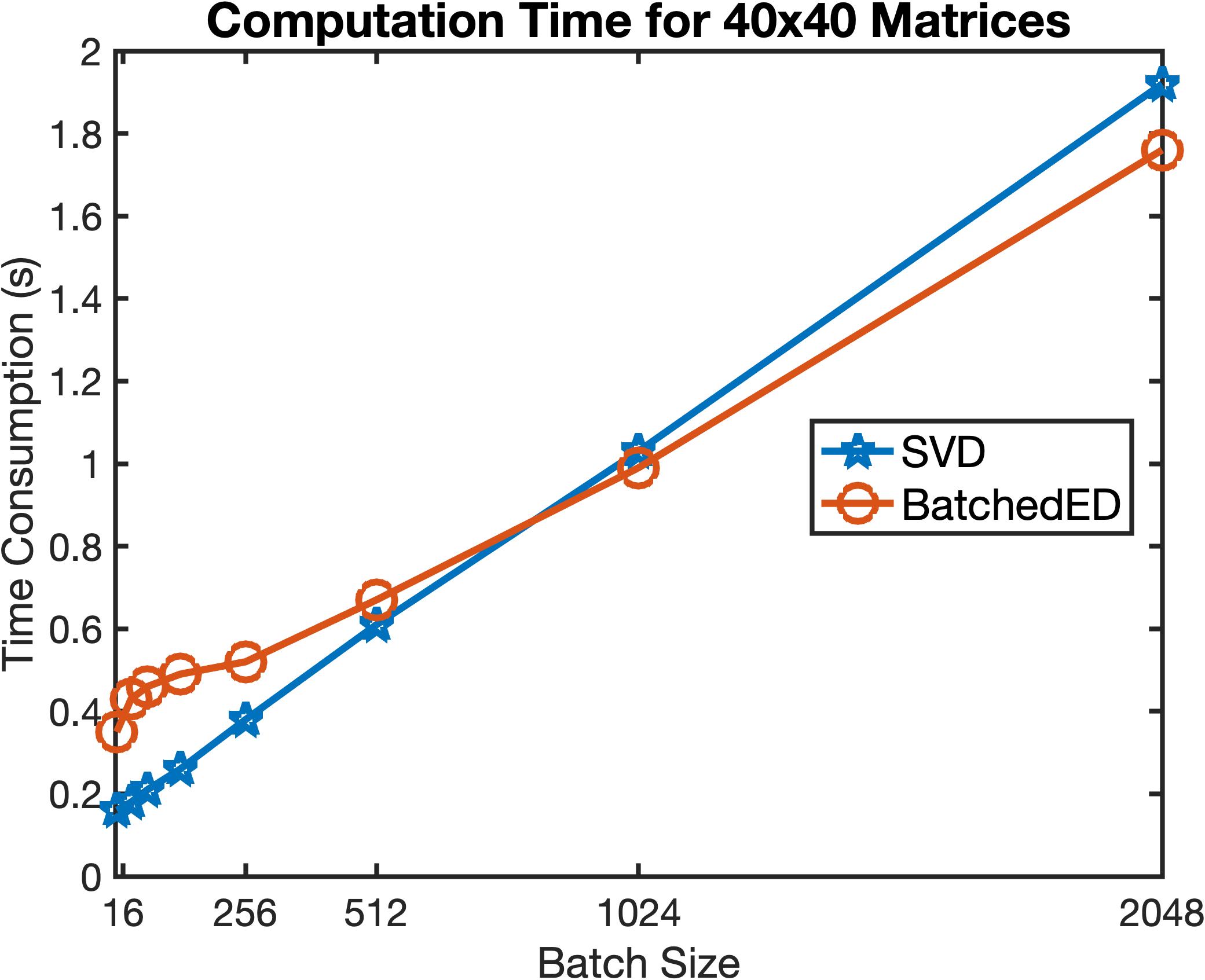}
    \caption{Time consumption for a mini-batch of $40{\times}40$ matrices. }
    \label{fig:batch_40}
\end{figure}

As discussed in the paper, our Batched ED fully utilizes the power of GPUs and can be very fast against varying batch sizes for small and medium matrices. However, one accompanying limitation is the cubic time cost $O{(}n^3{)}$ to the matrix dimension, which constrains our method to be applicable only to small-sized and moderate-sized matrices. A interesting question is to investigate where the critical point of our method against SVD lies, \emph{i.e.,} from what matrix dimension on, our method is no longer competitive against the SVD. Fig.~\ref{fig:batch_40} presents the time comparison for a mini-batch of matrices in the size $40{\times}40$. Our method only has the marginal advantage over the SVD when the batch size is larger than $1024$. This concludes that when the matrix dimension is larger than $40$ and the batch size is small, \textsc{torch.svd} can be a drop-in replacement for our method.  

\begin{figure}
    \centering
    \includegraphics[width=0.6\linewidth]{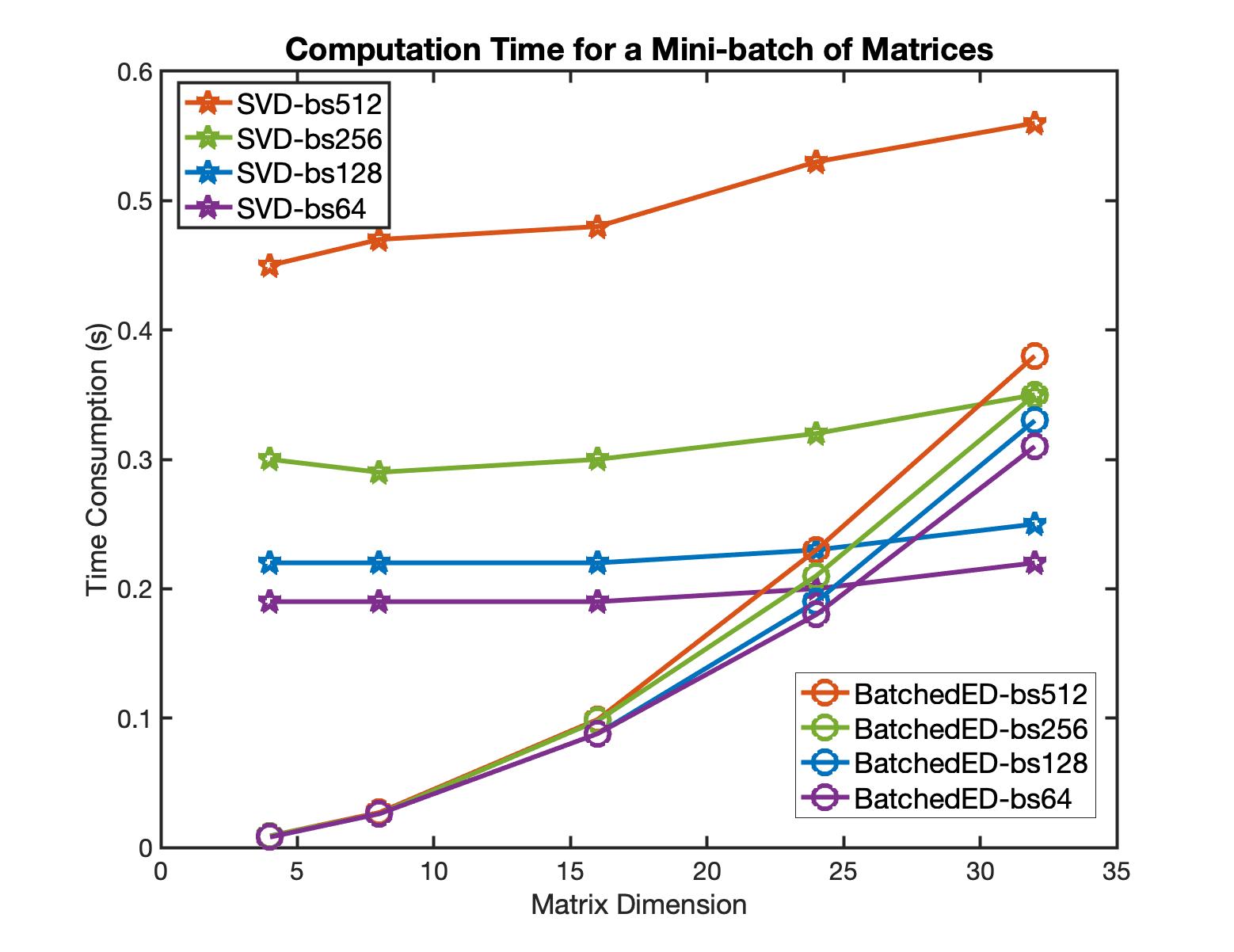}
    \caption{Time consumption of our Batched ED against SVD for a mini-batch of matrices in different batch sizes and matrix dimensions. Lines with the same batch sizes are in the same color.}
    \label{fig:svd_bed_bs}
\end{figure}

A similar question is when the batch size is fixed, in what range of matrix dimensions, our method holds a speed advantage over SVD. From the numerical test in the paper, we already know that when the matrix dimension is smaller than $24$, our Batched ED is consistently faster than the SVD. So the critical point should be larger than matrix dimension $24$. Fig.~\ref{fig:svd_bed_bs} compares the time consumption for batched matrices in batch sizes $512$, $256$, $128$, and $64$. The intersections of each pair of lines locate the critical points. For batch sizes $64$ and $128$, the intersection points are at matrix dimensions $26$ and $28$, respectively. When the batch size is $256$, our method has a faster speed for matrix dimensions less than $33$. As for matrices in batch size $512$,  the intersection might be around $40$. Larger batch sizes would make our Batched ED more advantageous.
\end{document}